\documentclass{article}
\usepackage{iclr2027_conference,times}
\iclrfinalcopy
\usepackage{amsmath}
\usepackage{amssymb}
\usepackage{hyperref}
\usepackage{url}

\usepackage{color}
\usepackage{xcolor}
\usepackage{cleveref}
\usepackage{caption}
\usepackage{amssymb}

\usepackage{booktabs}
\usepackage{multirow}
\usepackage{graphicx}
\usepackage{subcaption}
\usepackage{wrapfig}
\usepackage{tabularx}

\definecolor{green}{rgb}{0, 0.5, 0}
\definecolor{orange}{rgb}{0.6, 0.3, 0.1}
\definecolor{red}{rgb}{1.0, 0.0, 0.0}
\definecolor{teal}{rgb}{0.0, 0.4, 0.4}
\definecolor{purple}{rgb}{0.65,0,0.65}
\definecolor{saffron}{rgb}{0.95,0.75,0.2}
\definecolor{turquoise}{rgb}{0.0,0.5,0.5}
\definecolor{brown}{rgb}{0.5, 0.16, 0.16}
\definecolor{brickred}{rgb}{.6, .2, .1}
\definecolor{coral}{rgb}{1,0.45,0.33}
\definecolor{newcolor}{rgb}{.8,.349,.1}
\definecolor{ceruleanblue}{rgb}{0.16, 0.32, 0.75}
\definecolor{algcomment}{RGB}{75,115,116}

\newif\ifrevision
\revisionfalse       

\ifrevision

    \newcommand{\lmc}[1]{{\color{cyan}[LM: #1]}}

    \newcommand{\zsc}[1]{{\color{teal}[ZS: #1]}}
    \newcommand{\zs}[1]{{\color{blue} #1}}

    \newcommand{\dc}[1]{{\color{red} #1}}
    \newcommand{\dcc}[1]{{\color{red}[DC: #1]}}

    \newcommand{\dlc}[1]{{\color{orange}[DL: #1]}}

    \newcommand{\gz}[1]{{\color{purple} #1}}
    \newcommand{\gzc}[1]{{\color{purple}[GZ: #1]}}

    \newcommand{\hrc}[1]{{\color{ceruleanblue}[HR: #1]}}

\else

    \newcommand{\zs}[1]{#1}
    \newcommand{\dc}[1]{#1}
    
    \newcommand{\gz}[1]{#1}

    \newcommand{\lmc}[1]{}
    \newcommand{\zsc}[1]{}
    \newcommand{\dcc}[1]{}
    \newcommand{\dlc}[1]{}
    \newcommand{\gzc}[1]{}
    \newcommand{\hrc}[1]{}

\fi

\newcommand{\shortcite}[1]{\citeyearpar{#1}}
\newcommand{\method}{IntrinSync}

\title{IntrinSync: Joint Intrinsic Decomposition and Reciprocal Rendering}

\author{%
\parbox{\dimexpr\textwidth-2\tabcolsep\relax}{%
\centering
\normalfont
\textbf{Zheng Gu}$^{1}$ \hspace{1.5em}
\textbf{Rui Huang}$^{1}$ \hspace{1.5em}
\textbf{Xilu Zhang}$^{1}$ \hspace{1.5em}
\textbf{Jingbo Zhang}$^{2}$ \hspace{1.5em}
\textbf{Min Lu}$^{1}$ \\
\textbf{Zhida Sun}$^{1}$ \hspace{1.5em}
\textbf{Dani Lischinski}$^{3}$ \hspace{1.5em}
\textbf{Daniel Cohen-Or}$^{4}$ \hspace{1.5em}
\textbf{Hui Huang}$^{1}$\thanks{Corresponding author.}
\\[0.3em]
{\small
$^{1}$Shenzhen University
\hspace{1.5em}
$^{2}$Robotics X Lab, Tencent
\\
$^{3}$Hebrew University of Jerusalem
\hspace{1.5em}
$^{4}$Tel Aviv University
\endgraf
}%
}%
}

\usepackage{etoolbox}
\makeatletter
\patchcmd{\@maketitle}
  {\@title\par}
  {\@title\par\vspace{0.8em}}
  {}
  {\PackageWarning{arxiv-spacing}{Title spacing patch failed}}
\patchcmd{\@maketitle}
  {\vskip 0.3in minus 0.1in}
  {\vskip 0.1in}
  {}
  {\PackageWarning{arxiv-spacing}{Bottom spacing patch failed}}
\makeatother

\begin{document}
\maketitle
\lhead{Preprint}

\begin{center}
    \includegraphics[width=\textwidth]{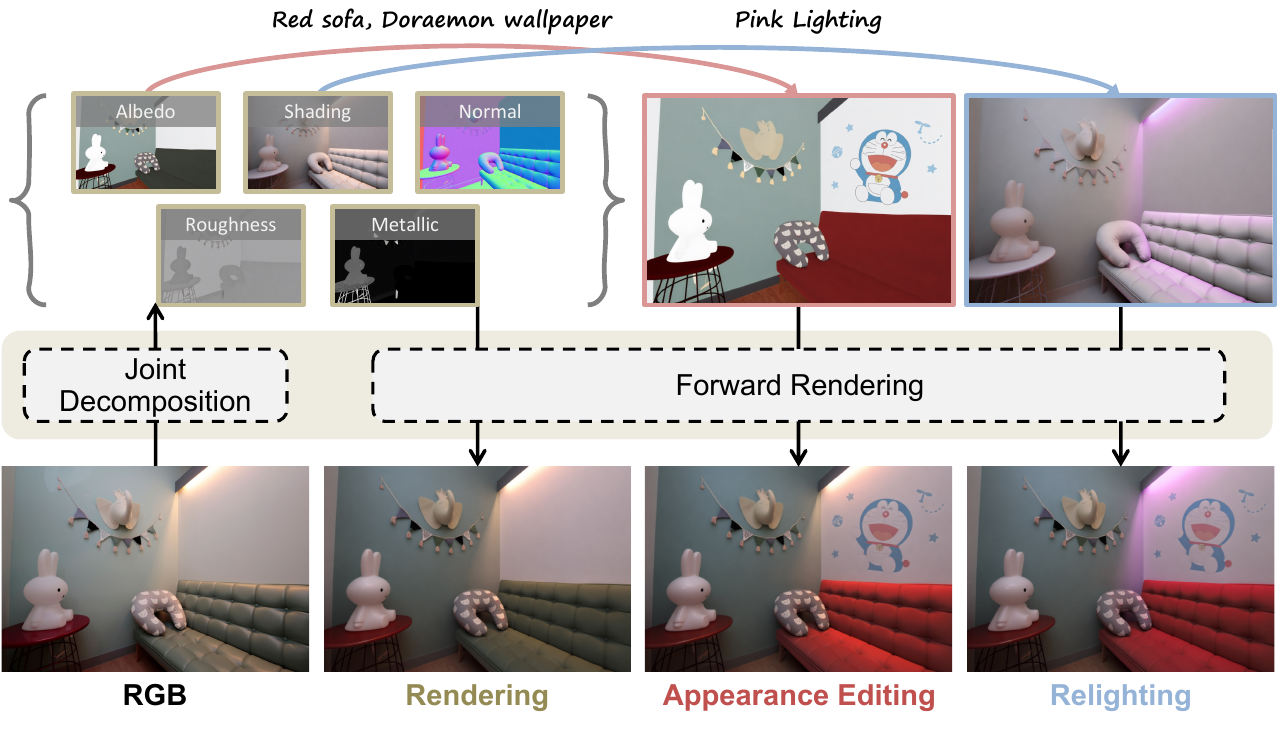}
    \captionof{figure}{
        \method{} jointly decomposes an input image into mutually dependent intrinsic components, enabling cross-channel information exchange and more coherent scene representations, while reciprocal forward rendering supports diverse applications such as material editing and relighting.
    }
    \label{fig:teaser}
\end{center}

\begin{abstract}
Inverse rendering decomposes an image into intrinsic properties such as appearance, illumination, geometry, and material, yet these properties are inherently interdependent.
A reliable decomposition should produce intrinsic maps that are not only individually plausible, but also mutually compatible in explaining the image.
\gz{However, existing methods either model intrinsic channels in isolation or treat inverse and forward rendering as separate processes, leaving the interdependence underexploited.}
In this paper, \gz{we introduce \method, a unified framework that captures this interdependence through joint-channel modeling and reciprocal inverse--forward rendering.}
At the channel level, we jointly decompose an input RGB into albedo, shading, surface normal, roughness, and metallic maps through a 1-to-N mapping, enabling information exchange across channels throughout generation.
At the process level, we \gz{establish inverse--forward reciprocity} through a dual cycle-consistent objective that aligns corresponding predictions across a closed loop.
Experiments on three datasets demonstrate that our method achieves competitive intrinsic estimation and forward rendering performance, improving coherence and physical consistency.
\gz{Beyond decomposition, \method{} provides a physically grounded interface for image editing, allowing intrinsic properties to be explicitly manipulated and rendered back into RGB images.}
\end{abstract}

\section{Introduction}

An RGB image of a scene entangles surface color, illumination, geometry, and material properties. 
Recovering these factors from a single image, commonly known as intrinsic image decomposition or inverse rendering, is a long-standing problem in computer vision and graphics~\citep{barrow1978recovering}. 
\dc{A fundamental challenge of this problem is that these intrinsic properties are interdependent--different combinations of these intrinsic properties can explain the same observed appearance.
}

\begin{figure*}[!t] 
\centering 
\includegraphics[width=\textwidth]{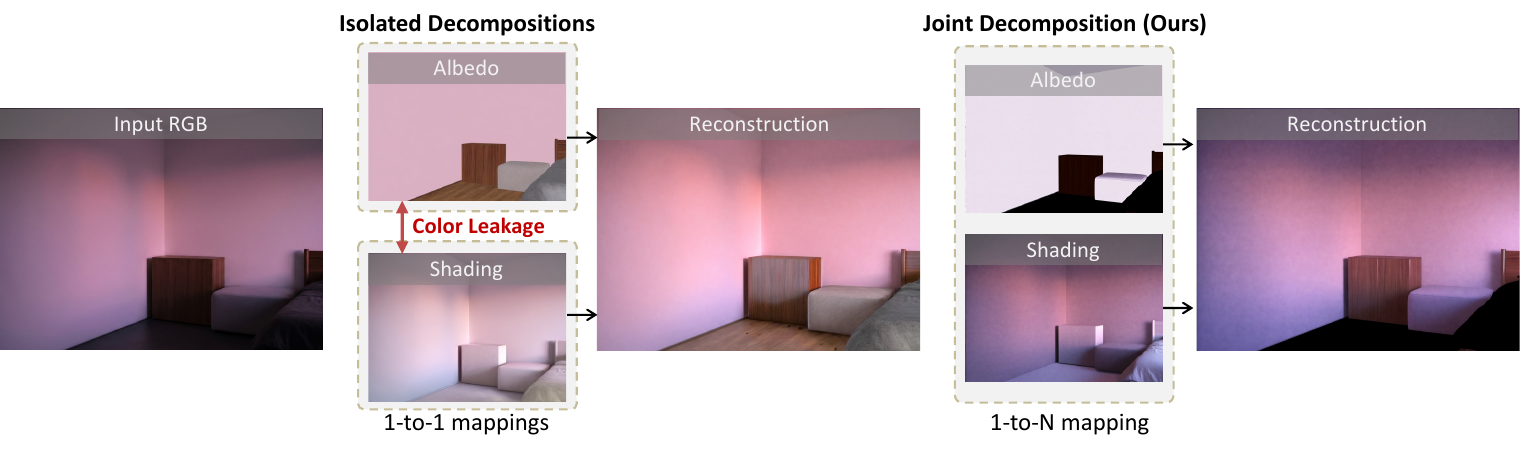} 
\caption{
\dc{\textbf{Joint intrinsic decomposition.}}
Traditional approaches predict intrinsic channels independently, leading to physically inconsistent estimations such as appearance leaking between intrinsic channels. 
\dc{Our joint 1-to-N decomposition allows information to be exchanged across channels, encouraging mutually coherent intrinsic maps that better explain the observed RGB image.}
} 
\label{fig:motivation} 
\end{figure*}

\gz{Such interdependence could be captured through two levels of interaction: among the intrinsic channels themselves, and between decomposition and its reverse process--forward rendering.}
Recent diffusion-based approaches have made substantial progress in modeling these intrinsic channels~\citep{zeng2024rgb,liang2025diffusion}. However, these two levels of interaction are explored separately.
Methods that jointly model intrinsic channels~\citep{luo2024intrinsicdiffusion,dirik2026prism} leave decomposition and rendering uncoupled, while those that close the loop between the two processes~\citep{sun2025ouroboros,zheng2025dnf} still estimate each intrinsic channel independently. 
Consequently, the predicted channels may individually appear plausible or even reconstruct the input well, yet still remain mutually inconsistent~(\Cref{fig:motivation}).
Therefore, we argue that intrinsic maps should be learned with both cross-channel interaction promoting mutual consistency, and reciprocal rendering providing feedback on how the intrinsic maps jointly explain the observed image.

\dc{In this paper, we \gz{turn this argument into} a unified framework, \gz{named \method}, \gz{which combines cross-channel interaction with cross-process reciprocity for intrinsic decomposition and forward rendering.}}
We encode an RGB image and its intrinsic maps as a joint sequence of tokens within a Multimodal Diffusion Transformer (MMDiT)~\citep{wu2025qwenimagelayered}. 
For inverse rendering, the model jointly denoises all intrinsic tokens conditioned on the RGB image, producing albedo, shading, surface normal, roughness, and metallic maps through a 1-to-N mapping. 
Attention across these tokens allows each channel to use information from the others during generation. 
For forward rendering, the same architecture conditions on the intrinsic maps to synthesize the corresponding RGB image through an N-to-1 mapping.

To further couple the joint intrinsic representation with the forward process, we introduce a \gz{dual cycle-consistent objective}. 
We construct two reciprocal paths: one decomposes an RGB image and renders it back, while the other renders intrinsic maps and decomposes the resulting image. 
In addition to direct supervision, the cycle objective aligns corresponding predictions across these paths, allowing the forward and inverse processes to mutually constrain the learned scene representation.
To improve generalization to real-world scenes, we also curate indoor and outdoor photographs along with synthetic training data. 
This allows the model to learn from real-world appearances without ground-truth intrinsic annotations.

We evaluate our method on large-scale benchmarks HyperSim~\citep{roberts2021hypersim}, InteriorVerse~\citep{zhu2022learning}, and MatrixCity~\citep{li2023matrixcity}, covering both indoor and outdoor scenes. 
Across intrinsic estimation and forward rendering benchmarks, our unified model achieves competitive or improved performance while producing \dc{more coherent and physically consistent intrinsic decompositions.}
\gz{Beyond decomposition, our reciprocal formulation enables intrinsic-guided image editing~(\Cref{fig:teaser}). We can first decompose an input image into editable intrinsic maps, manipulate the material, illumination, or geometry properties, and render them back into RGB through \method. This enables fine-grained image manipulation through explicit control of physical properties, while preserving the content of the scene.}

\section{Related Work}

\subsection{Image Diffusion Models}

Diffusion models~\citep{ho2020denoising,flux2024} have enabled high-quality image synthesis and have also been adapted to image decomposition. 
For semantic concept decomposition, Break-A-Scene~\citep{avrahami2023break} and Inspiration-Tree~\citep{vinker2023concept} learn representations of visual concepts for reuse and exploration, while UnZipLoRA~\citep{liu2025unziplora} separates subject and style.
For layered image decomposition, previous approaches explore transparent layer generation~\citep{yang2025generative} and image decomposition into editable RGBA components~\citep{zhang2024transparent}, alongside work on layered vector representations~\citep{wang2025layered,wu2025layerpeeler}. Qwen-Image-Layered~\citep{wu2025qwenimagelayered} jointly decomposes an input image into multiple RGBA layers through inter-layer attention within an MMDiT. 
Building upon its multilayer modeling capability, we target physical scene properties rather than semantic or compositing layers, and learn their estimation together with forward rendering. \gz{This enables interactions among intrinsic properties within a unified representation.}

\subsection{Intrinsic Decomposition}

Early optimization-based methods constrain intrinsic decomposition using priors on shading and reflectance~\citep{barron2012shape,barron2014shape}. Additional observations such as depth and multiple views~\citep{philip2019multi,ye2023intrinsicnerf} have also been used to reduce ambiguity. 
With the availability of large-scale annotated synthetic datasets, data-driven intrinsic estimation approaches have been developed subsequently. Datasets such as Hypersim~\citep{roberts2021hypersim}, InteriorVerse~\citep{zhu2022learning}, and MatrixCity~\citep{li2023matrixcity} provide dense annotations for various \zs{scene properties}. Neural approaches learn decomposition from data~\citep{wang2023discriminative,liu2020unsupervised}. \citet{careaga2023intrinsic} introduces ordinal shading estimation, while the subsequent work~\citep{careaga2024colorful} extends the decomposition to colored diffuse shading.

More recently, the generative priors of pretrained diffusion models have offered a powerful alternative for intrinsic decomposition. RGB$\leftrightarrow$X~\citep{zeng2024rgb} uses separate models for per-channel intrinsic prediction and image synthesis from intrinsic maps. Ouroboros~\citep{sun2025ouroboros} fine-tunes inverse and forward models with bidirectional cycle losses, while DNF-Intrinsic~\citep{zheng2025dnf} learns deterministic image-to-intrinsic mappings. IntrinsicDiffusion~\citep{luo2024intrinsicdiffusion} and PRISM~\citep{dirik2026prism} learn multiple intrinsic modalities through shared representation and modality-specific prompts. 
ReasonX~\citep{dirik2026reasonx} uses relative intrinsic judgments from a VLM as rewards to fine-tune existing predictors. Some frontier work have also extended the intrinsic properties to the video domain~\citep{liang2025diffusion,chen2026unividx}.

\gz{Different from prior work, we explicitly model the interdependence of intrinsic properties at two complementary levels within a unified reciprocal rendering framework. 
(a) At the channel level, we jointly denoise all intrinsic channels within a shared contextual sequence, allowing each channel to exchange information with the others.
(b) At the process level, instead of treating decomposition and rendering as isolated tasks, we couple them within a unified framework. 
Together, these reciprocal processes provide consistency feedback in addition to direct supervision.
}

\section{Method}
In this section, we introduce our method for joint intrinsic decomposition and reciprocal forward rendering. We clarify our problem formulation in \Cref{sec:problem_formulation}, illustrate our unified framework for channel-level joint modeling of intrinsic maps in \Cref{sec:framework}, and introduce our process level reciprocal rendering with dual cycle-consistency in \Cref{sec:training_strategy}.



\subsection{Problem Formulation}
\label{sec:problem_formulation}

\gz{We study inverse and forward rendering between an RGB image $I \in \mathbb{R}^{H\times W\times 3}$ and its intrinsic maps $\mathcal{X} = \{A, S, N, R, M\}$. Here, $A, S, N \in \mathbb{R}^{H\times W\times 3}$ denote albedo, shading, and camera-space surface normals, respectively, while $R, M \in \mathbb{R}^{H\times W\times 1}$ denote roughness and metallic maps.
All intrinsic maps are spatially aligned with the RGB image. Inverse rendering estimates the full set $\mathcal{X}$ from $I$, whereas forward rendering synthesizes an RGB image conditioned on the intrinsic maps.
Our goal is to learn both directions within a unified framework, while encouraging the predicted intrinsic maps to form a consistent explanation for image reconstruction.}

\begin{figure*}[!t]
    \centering
    \includegraphics[width=\linewidth]{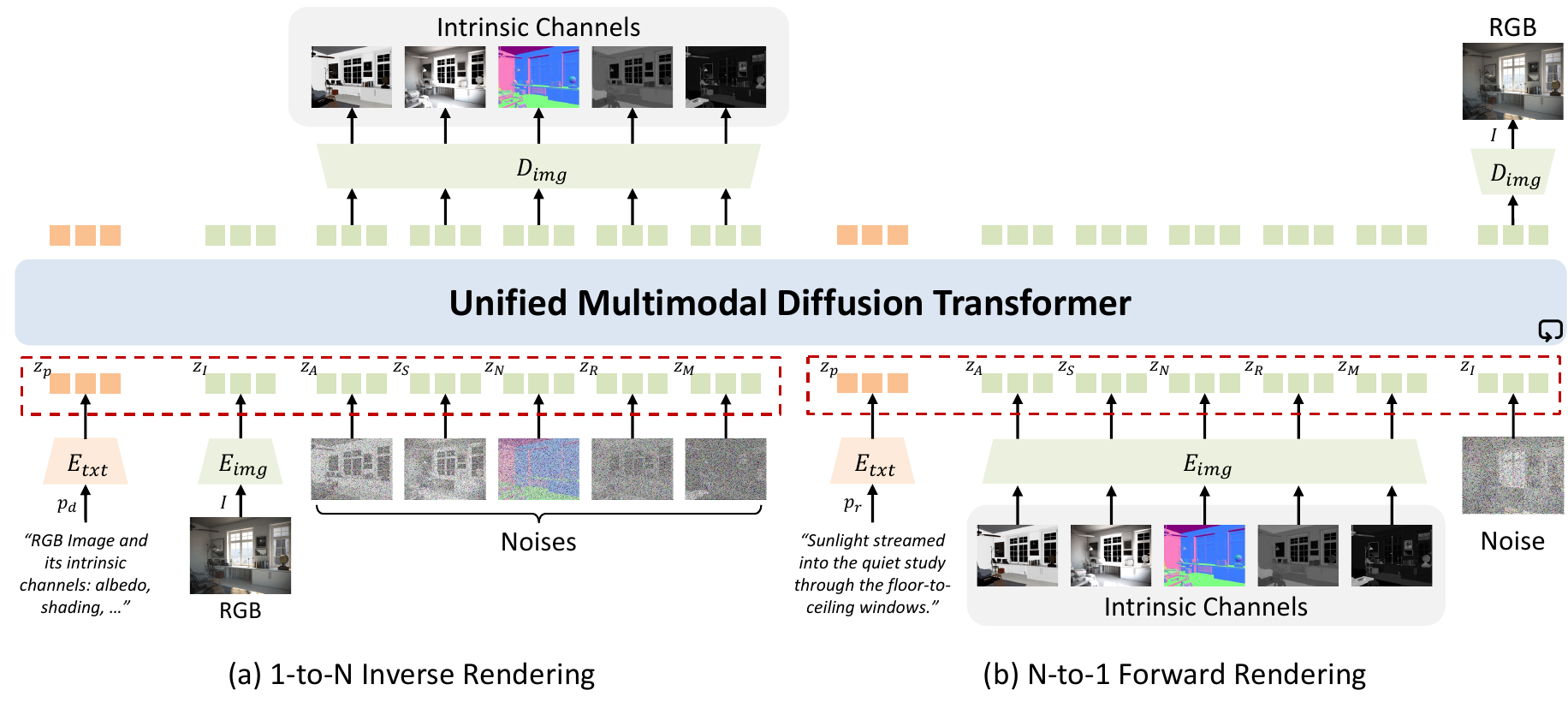}
    \caption{
    \textbf{Overview of joint-channel intrinsic decomposition with reciprocal rendering.}
    For inverse rendering, the input RGB image conditions a joint 1-to-N denoising process that predicts multiple intrinsic channels simultaneously.
    For forward rendering, the predicted intrinsic channels jointly condition an N-to-1 process to synthesize the corresponding RGB image.
    }
    \label{fig:framework}
\end{figure*}

\subsection{Joint-channel Intrinsic Modeling}
\label{sec:framework}

We build our model on Qwen-Image-Layered~\citep{wu2025qwenimagelayered}, retaining its 3D RoPE~\citep{su2024roformer} to represent spatial and layer positions. We organize the intrinsic channel tokens in a fixed order, which helps to preserve the spatial alignment. 
Inverse rendering jointly generates intrinsic maps conditioned on the RGB image, while forward rendering generates RGB conditioned on the intrinsic maps. 
We use the image encoder $E_{\mathrm{img}}$ to encode the RGB image and each intrinsic map into latent space first. For roughness and metallic maps, we expand the input into three channels before encoding. 
Then, the visual latents are patchified into token sequences.
Each intrinsic component retains its own latent representation, while attention allows its features to interact with those of the other components. 

\paragraph{Inverse Rendering}
For inverse rendering, we estimate the full intrinsic set
$\mathcal{X}$ from an RGB image $I$. Per-channel approaches such as RGB$\leftrightarrow$X~\citep{zeng2024rgb} and Ouroboros~\citep{sun2025ouroboros} generate each intrinsic map in an isolated denoising process with a channel-specific prompt $p$ as a ``switch'' to specify the target. 
Such a formulation treats the generated channels as conditionally independent. Each channel is sampled without reference to the interpretations of the others, yielding incompatible explanations of the same RGB observation.

In contrast, we model the joint conditional distribution $p_{\Phi}(\mathcal{X}\mid I,p_d)$ directly, without imposing the channel-wise factorization. Here $p_d$ denotes a fixed prompt specifying all of the target intrinsic channels. As shown in \Cref{fig:framework}, during decomposition, the target latents are initialized from Gaussian noise and denoised together. Therefore, the update of each channel can depend on the evolving features of the entire intrinsic set, allowing cross-channel dependencies captured throughout the generation process. The final latents are decoded by $D_{\mathrm{img}}$ into the five intrinsic maps.

\paragraph{Forward Rendering}

For forward rendering, we exchange the roles of conditions and targets, modeling the conditional distribution $p_{\Phi}(I\mid\mathcal{X},p_r)$. As shown in \Cref{fig:framework}, the rendering process uses the same representation and backbone. While prior works also explore the X$\rightarrow$RGB problem under a diffusion framework, our approach is unique \zs{in} that we employ the exact same architecture used for the inverse task without any architecture modification. This reciprocity narrows the gap between inverse and forward rendering.
We retain all five conditioning slots and fill unavailable channels with zeros, allowing training on heterogeneous datasets with different annotations. This N-to-1 mapping connects the \zs{jointly} predicted intrinsic maps back to the RGB domain, forming the reciprocal paths in our cycle consistency objective.

\subsection{Dual Cycle-Consistent Training}
\label{sec:training_strategy}

\paragraph{Direct supervision}
Following the rectified flow formulation, given paired data consisting of an RGB image $I$ and its intrinsic channels $\mathcal{X}$, we first define a supervised reconstruction loss $\mathcal{L}_{rec}$ to learn the bidirectional mappings:
\begin{equation}
\mathcal{L}_{rec} = \mathbb{E}_{t} \left[ || v_\Phi(z_{\mathcal{X}}^{t}, t, z_I, p_d) - v_{\mathcal{X}} ||^2 + || v_\Phi(z_{I}^{t}, t, z_{\mathcal{X}}, p_r) - v_I ||^2 \right],
\end{equation}
where $t$ denotes timestep, $\{z_{\mathcal{X}}^{t}, z_{I}^{t}\}$ are noised latents, $\{z_I, z_\mathcal{X}\}$ are clean latents, and $\{v_\mathcal{X},v_I\}$ denote target velocity. This supervision makes both directions learn the correct mapping.
However, the two processes are still optimized separately. The decomposition output is not directly checked by the rendering process, and vice versa.

\begin{figure*}[!t] 
\centering \includegraphics[width=\textwidth]{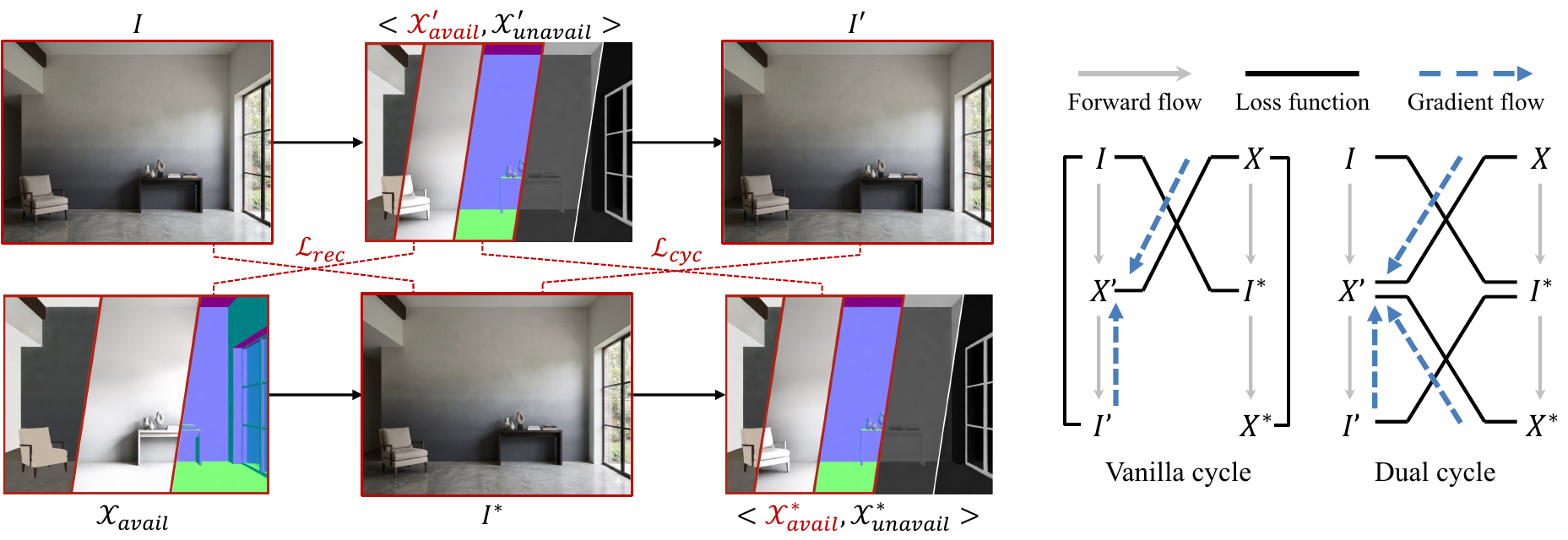} \caption{\textbf{Dual cycle-consistent training.} We couple decomposition and rendering through two reciprocal paths.
The vanilla cycle constrains only endpoint reconstruction, whereas our dual cycle aligns corresponding predictions across the two paths, enforcing process-level reciprocity.
Direct supervision is applied to available intrinsic channels for heterogeneous training data.} 
\label{fig:cycle} 
\end{figure*}

\paragraph{Dual Cycle-Consistent Tuning}
A natural way to connect the two directions is introducing a cycle consistency loss. As illustrated in Figure~\ref{fig:cycle}, we construct two reciprocal paths: $I \rightarrow \mathcal{X}' \rightarrow I'$ and $\mathcal{X} \rightarrow I^* \rightarrow \mathcal{X}^*$. A vanilla cycle enforces $I' \approx I$ and $\mathcal{X}^* \approx \mathcal{X}$. This constrains the final output of the composed mappings, but does not directly constrain the intermediate prediction. Thus, the model may find shortcuts, reconstruct the input well even if the predicted intrinsic maps are inaccurate.

Instead of only matching the cycle endpoints, we constrain the corresponding prediction processes along the two reciprocal paths. For decomposition, the prediction from the rendered image $I^*$ is aligned with the directly supervised decomposition from $I$. For rendering, the prediction conditioned on $\mathcal{X}'$ is aligned with the directly supervised rendering from $\mathcal{X}$:
\begin{equation}
\begin{split}
\mathcal{L}_{cyc} = \mathbb{E}_{t_1} \left[ || v_\Phi(z_{\mathcal{X^*}}^{t_1}, t_1, z_{I^*},p_d) - v_\Phi(z_\mathcal{X}^{t_1},t_1,z_I,p_d) || \right] \\
+ \mathbb{E}_{t_2} \left[ || v_\Phi(z_{I'}^{t_2}, t_2, z_{\mathcal{X}'},p_r) - v_\Phi(z_I^{t_2},t_2,z_\mathcal{X},p_r) || \right],
\end{split}
\end{equation}
where $\{t_1,t_2\}$ are two different timesteps. Through this dual cycle design, the model is encouraged to keep the same decomposition and rendering behavior after crossing to the other process. We jointly optimize inverse and forward rendering using a combination of the direct supervision and dual cycle-consistency losses in our training.

\begin{figure*}[!t] 
\centering \includegraphics[width=\textwidth]{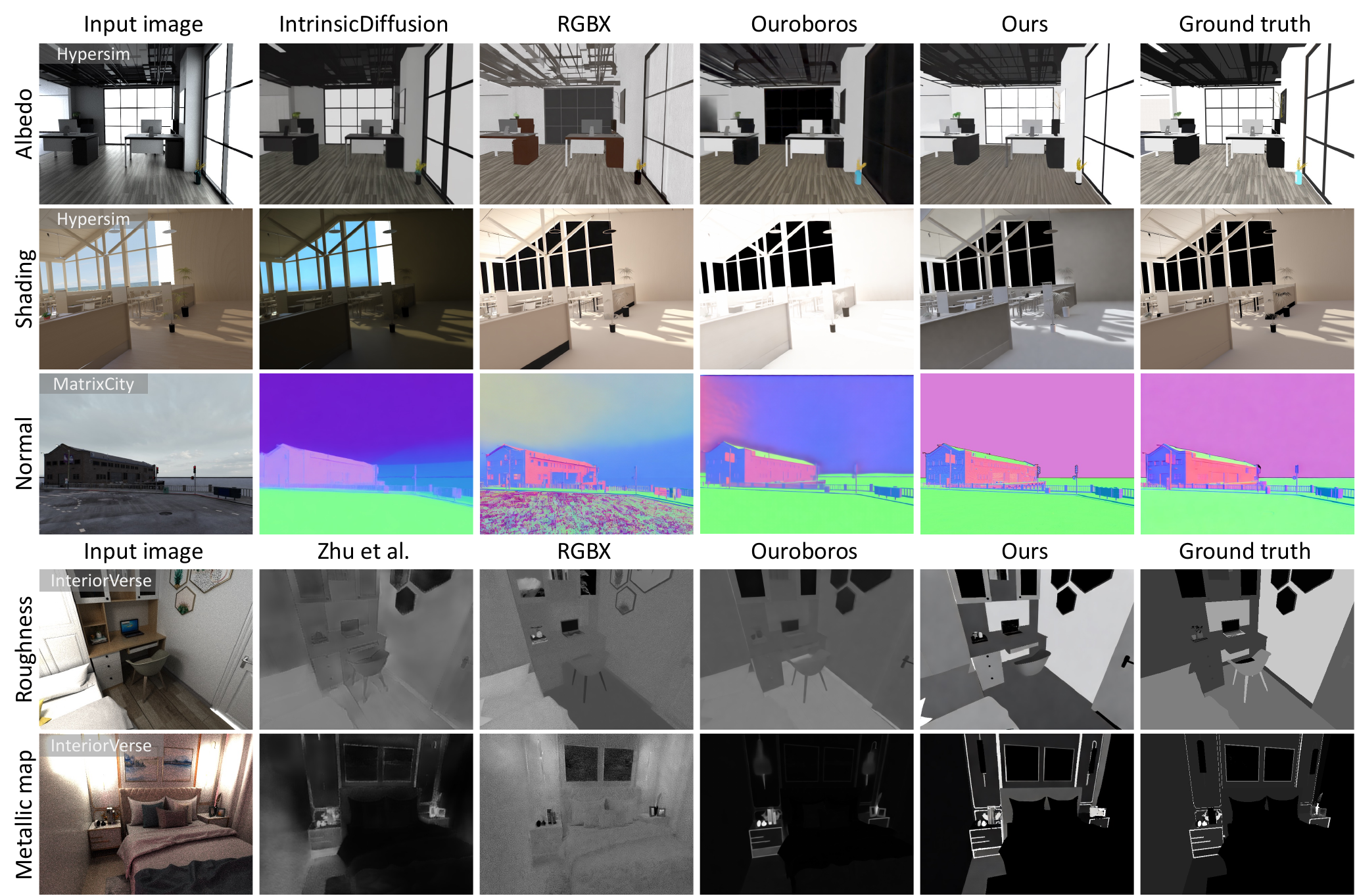} \caption{\textbf{Qualitative Comparison of inverse rendering}. We visualize qualitative comparison with baseline methods on Hypersim~\citep{roberts2021hypersim}, InteriorVerse~\citep{zhu2022learning}, and MatrixCity~\citep{li2023matrixcity}. Each row shows an input image and decomposed intrinsic channels from different methods. More results are shown in the appendix.} 
\label{fig:qualitative_comparison} 
\end{figure*} 

\section{Experiments}

\subsection{Implementation}
We curate our training data by collecting 20K images from Hypersim~\citep{roberts2021hypersim}, 15K images from InteriorVerse~\citep{zhu2022learning}, and 20K images from MatrixCity~\citep{li2023matrixcity}, making a total of 55K $\left \langle I, \mathcal{X}_{avail} \right \rangle $ pairs.
Following previous work~\citep{fang2026vrgbx}, we also leverage real-world RGB images to improve generalization to in-the-wild images. Specifically, we include 45K real-world images from MIDIntrinsics~\citep{murmann2019multi}, ScanNet++~\citep{yeshwanth2023scannetpp}, DL3DV~\citep{ling2024dl3dv}, and Adobe FiveK~\citep{vladimir2011fivek}.
We use a joint and fixed prompt template $p_d$ for decomposition as ``\textit{Albedo, Shading, Camera-space Normal, Roughness, Metallic}'', and utilize Qwen-VL~\citep{wu2025qwen} to obtain image captions $p_r$ of the RGB images. See the appendix for further details.

\subsection{Evaluation Setup}



We evaluate our method on test subsets of Hypersim~\citep{roberts2021hypersim}, InteriorVerse~\citep{zhu2022learning}, and MatrixCity~\citep{li2023matrixcity}.
We compare our approach to Zhu et al.~\shortcite{zhu2022learning}, Careaga et al.~\shortcite{careaga2024colorful},  RGB$\leftrightarrow$X~\citep{zeng2024rgb}, DiffusionRenderer~\citep{liang2025diffusion}, IntrinsicDiffusion~\citep{luo2024intrinsicdiffusion}, and Ouroboros~\citep{sun2025ouroboros}. We use their officially provided checkpoints of these baselines and run inference on the same test data.
To evaluate the intrinsic maps predicted for inverse rendering, we adopt PSNR and LPIPS metrics. Furthermore, Root Mean Square Error (RMSE) and SSIM are utilized specifically for albedo estimation. For normal estimation, we measure the mean angular error and the percentage of pixels with an angular error below $11.25^\circ$. For forward rendering and RGB reconstruction, we assess the overall reconstruction quality using PSNR, LPIPS, and SSIM.\footnote{Forward rendering reconstructs the RGB image from ground-truth intrinsic maps, whereas RGB reconstruction uses predicted intrinsic maps to render them back to RGB.}

\begin{table*}[t]
    \centering
    \caption{
    \textbf{Quantitative comparison with baseline methods on synthetic datasets.}
    The best and second-best results are highlighted in
    \textbf{bold} and \underline{underlined}, respectively.
    ``--'' denotes unavailable results.
    }
    \label{tab:quantitative_comparison}
    \setlength{\tabcolsep}{2.5pt}
    \renewcommand{\arraystretch}{1.15}
    \resizebox{\textwidth}{!}{%
    \begin{tabular}{lcccccccccccccc}
        \toprule
        \multirow{2}{*}{Method}
        & \multicolumn{4}{c}{Albedo}
        & \multicolumn{2}{c}{Normal}
        & \multicolumn{2}{c}{Roughness}
        & \multicolumn{2}{c}{Metallic}
        & \multicolumn{2}{c}{Fwd. Rendering}
        & \multicolumn{2}{c}{RGB Recon.} \\
        \cmidrule(lr){2-5}
        \cmidrule(lr){6-7}
        \cmidrule(lr){8-9}
        \cmidrule(lr){10-11}
        \cmidrule(lr){12-13}
        \cmidrule(lr){14-15}
        & PSNR$\uparrow$ & LPIPS$\downarrow$
        & SSIM$\uparrow$ & RMSE$\downarrow$
        & Mean$\downarrow$ & $11.25^\circ\uparrow$
        & PSNR$\uparrow$ & LPIPS$\downarrow$
        & PSNR$\uparrow$ & LPIPS$\downarrow$
        & PSNR$\uparrow$ & LPIPS$\downarrow$
        & PSNR$\uparrow$ & LPIPS$\downarrow$ \\
        \midrule

        \multicolumn{15}{c}{{\textbf{Hypersim}}} \\
        \midrule

        Zhu et al.
        & 17.360 & 0.517 & 0.542 & 0.165
        & 25.079 & 46.69
        & -- & --
        & -- & --
        & -- & --
        & -- & -- \\

        Careaga et al.
        & 18.131 & 0.321 & \underline{0.709} & 0.137
        & -- & --
        & -- & --
        & -- & --
        & -- & --
        & -- & -- \\

        IntrinsicDiffusion
        & 19.264 & 0.336 & 0.641 & 0.131
        & 22.089 & 48.90
        & -- & --
        & -- & --
        & -- & --
        & -- & -- \\

        DiffusionRenderer
        & 19.919 & 0.344 & 0.661 & 0.124
        & {16.465} & {71.75}
        & -- & --
        & -- & --
        & -- & --
        & -- & -- \\

        RGB$\leftrightarrow$X
        & 18.921 & \underline{0.223} & 0.641 & 0.122
        & \underline{14.167} & \underline{76.51}
        & -- & --
        & -- & --
        & 15.364 & \underline{0.341}
        & \textbf{17.635} & \underline{0.236} \\

        Ouroboros
        & \underline{21.511} & 0.291 & 0.684 & \underline{0.101}
        & \textbf{13.293} & \textbf{78.77}
        & -- & --
        & -- & --
        & \textbf{17.968} & 0.346
        & 16.726 & 0.373 \\

        \textbf{\method{} (Ours)}
        & \textbf{22.445} & \textbf{0.215}
        & \textbf{0.779} & \textbf{0.090}
        & 18.758 & 48.48
        & -- & --
        & -- & --
        & \underline{17.235} & \textbf{0.228}
        & \underline{17.148} & \textbf{0.236} \\

        \midrule
        \multicolumn{15}{c}{{\textbf{InteriorVerse}}} \\
        \midrule

        Zhu et al.
        & 12.106 & 0.439 & 0.626 & 0.262
        & 12.487 & 74.02
        & 14.127 & 0.410
        & 14.868 & 0.406
        & -- & --
        & -- & -- \\

        Careaga et al.
        & 11.737 & 0.468 & 0.544 & 0.273
        & -- & --
        & -- & --
        & -- & --
        & -- & --
        & -- & -- \\

        IntrinsicDiffusion
        & 13.345 & 0.317 & 0.679 & 0.229
        & 14.895 & 64.15
        & -- & --
        & -- & --
        & -- & --
        & -- & -- \\

        DiffusionRenderer
        & 12.108 & 0.527 & 0.484 & 0.273
        & 11.701 & 79.02
        & 11.033 & 0.467
        & 13.558 & 0.493
        & -- & --
        & -- & -- \\

        RGB$\leftrightarrow$X
        & 12.681 & 0.523 & 0.505 & 0.245
        & 25.002 & 63.26
        & 11.331 & 0.674
        & 7.252 & 0.800
        & 10.180 & \underline{0.625}
        & 12.307 & \underline{0.413} \\

        Ouroboros
        & \underline{16.345} & \underline{0.246}
        & \underline{0.778} & \underline{0.171}
        & \textbf{8.714} & \textbf{84.73}
        & \textbf{16.202} & \underline{0.325}
        & \textbf{19.740} & \textbf{0.269}
        & \underline{11.256} & 0.656
        & \underline{12.780} & 0.579 \\

        \textbf{\method{} (Ours)}
        & \textbf{16.408} & \textbf{0.227}
        & \textbf{0.783} & \textbf{0.164}
        & \underline{11.190} & \underline{79.82}
        & \underline{14.947} & \textbf{0.299}
        & \underline{17.184} & \underline{0.272}
        & \textbf{12.637} & \textbf{0.444}
        & \textbf{16.636} & \textbf{0.350} \\

        \midrule
        \multicolumn{15}{c}{{\textbf{MatrixCity}}} \\
        \midrule

        Careaga et al.
        & 21.583 & \underline{0.324} & 0.371 & 0.095
        & -- & --
        & -- & --
        & -- & --
        & -- & --
        & -- & -- \\

        IntrinsicDiffusion
        & 18.681 & 0.535 & 0.332 & 0.135
        & \underline{22.359} & 46.39
        & -- & --
        & -- & --
        & -- & --
        & -- & -- \\

        DiffusionRenderer
        & 19.332 & 0.450 & 0.376 & 0.123
        & 24.355 & 48.26
        & 8.593 & 0.593
        & 13.578 & 0.429
        & -- & --
        & -- & -- \\

        RGB$\leftrightarrow$X
        & 20.115 & 0.364 & 0.451 & 0.112
        & 43.151 & 22.79
        & 7.839 & 0.585
        & 7.284 & 0.870
        & 9.733 & 0.501
        & 10.019 & \underline{0.462} \\

        Ouroboros
        & \underline{22.034} & 0.366
        & \underline{0.503} & \underline{0.088}
        & 23.803 & \underline{53.28}
        & \underline{12.012} & \underline{0.529}
        & \underline{19.218} & \underline{0.272}
        & \underline{17.566} & \underline{0.311}
        & \underline{13.187} & 0.516 \\

        \textbf{\method{} (Ours)}
        & \textbf{27.867} & \textbf{0.136}
        & \textbf{0.816} & \textbf{0.047}
        & \textbf{13.254} & \textbf{68.60}
        & \textbf{20.881} & \textbf{0.288}
        & \textbf{24.429} & \textbf{0.147}
        & \textbf{22.104} & \textbf{0.196}
        & \textbf{21.320} & \textbf{0.198} \\

        \bottomrule
    \end{tabular}%
    }
\end{table*}

\subsection{Qualitative Results}
We present qualitative comparison in \Cref{fig:qualitative_comparison} with baseline methods IntrinsicDiffusion~\citep{luo2024intrinsicdiffusion}, Zhu et al.~\shortcite{zhu2022learning}, RGB$\leftrightarrow$X~\citep{zeng2024rgb}, and Ouroboros~\citep{sun2025ouroboros} on multiple inverse rendering datasets and tasks. Our approach demonstrates competitive decomposition performance across diverse datasets and multiple tasks.
Notably, our method is capable of predicting intrinsic physical properties that are not finely annotated in the datasets. For instance, in the second row, our method recovers the illumination of unlabeled dark regions on the floor in the Hypersim dataset~\citep{roberts2021hypersim}. Another example is shown in the fifth row, where our method clearly reconstructs blurry metal connectors above and below the wall-mounted lamps. Additional results are shown in the appendix.

\subsection{Quantitative Results}
We quantitatively compare our method with state-of-the-art approaches on Hypersim~\citep{roberts2021hypersim}, InteriorVerse~\citep{zhu2022learning}, and MatrixCity~\citep{li2023matrixcity}. As summarized in \Cref{tab:quantitative_comparison}, our method achieves strong performance across diverse scene domains and intrinsic properties, while jointly supporting both inverse and forward rendering within the same framework.
For intrinsic decomposition, our method consistently improves albedo estimation across all three benchmarks, also achieving competitive results in normal, roughness, and metallic estimation, particularly on MatrixCity~\citep{li2023matrixcity}. These results suggest that jointly reasoning over multiple intrinsic channels is particularly beneficial when handling the diverse geometry, materials, and illumination found in complex scenes. 
Moreover, our method also performs strongly in the forward direction. These results demonstrate that the predicted intrinsic channels are not only individually accurate, but also form a mutually compatible scene representation that supports faithful reconstruction and neural rendering.

\begin{table}[t]
    \centering
    \setlength{\belowcaptionskip}{5pt}
    \caption{
        \textbf{Ablation study.}
        All the variants are trained on a subset of the training data from Hypersim.
    }
    \label{tab:ablation}

    \begingroup
    \small
    \setlength{\tabcolsep}{2pt}
    \renewcommand{\arraystretch}{1.15}

    \begin{tabular*}{\linewidth}{@{\extracolsep{\fill}}l*{10}{c}@{}}
        \toprule
        & \multicolumn{2}{c}{Albedo}
        & \multicolumn{2}{c}{Irradiance}
        & \multicolumn{2}{c}{Normal}
        & \multicolumn{2}{c}{Fwd. Rendering}
        & \multicolumn{2}{c}{RGB Recon.} \\

        \cmidrule(lr){2-3}
        \cmidrule(lr){4-5}
        \cmidrule(lr){6-7}
        \cmidrule(lr){8-9}
        \cmidrule(lr){10-11}

        & PSNR$\uparrow$ & LPIPS$\downarrow$
        & PSNR$\uparrow$ & LPIPS$\downarrow$
        & Mean$\downarrow$ & $11.25^\circ\!\uparrow$
        & PSNR$\uparrow$ & LPIPS$\downarrow$
        & PSNR$\uparrow$ & LPIPS$\downarrow$ \\
        \midrule

        (a) 1-to-1 decomp.
        & 17.915 & 0.285
        & 17.402 & 0.294
        & 25.712 & 37.74
        & --- & ---
        & --- & --- \\

        (b) N-to-1 rendering
        & --- & ---
        & --- & ---
        & --- & ---
        & 17.714 & 0.235
        & 16.664 & 0.288 \\

        (c) Joint decomp.
        & \underline{19.184} & 0.235
        & \underline{18.307} & 0.254
        & 23.390 & 41.23
        & --- & ---
        & --- & --- \\

        (d) Joint + vanilla cycle
        & 18.781 & \textbf{0.206}
        & 18.211 & \textbf{0.239}
        & \underline{19.365} & \underline{50.19}
        & \underline{18.025} & \underline{0.219}
        & \textbf{19.767} & \textbf{0.145} \\

        (e) Joint + dual cycle
        & \textbf{19.250} & \underline{0.229}
        & \textbf{18.633} & \underline{0.248}
        & \textbf{19.039} & \textbf{54.87}
        & \textbf{18.092} & \textbf{0.217}
        & \underline{18.132} & \underline{0.236} \\

        \bottomrule
    \end{tabular*}

    \endgroup
\end{table}

\subsection{Ablation Study}
We conduct \zs{an ablation study} to evaluate the contributions of the two key components. We use a subset of the whole training data for the ablation study.
Two baselines are trained for (a) prompt-conditioned 1-to-1 intrinsic decomposition and (b) N-to-1 forward rendering, without joint-channel modeling and cycle constraints.
Then, we train variants using the same data and evaluate them on the Hypersim subset.



\begin{figure}[!t] 
\centering \includegraphics[width=\linewidth]{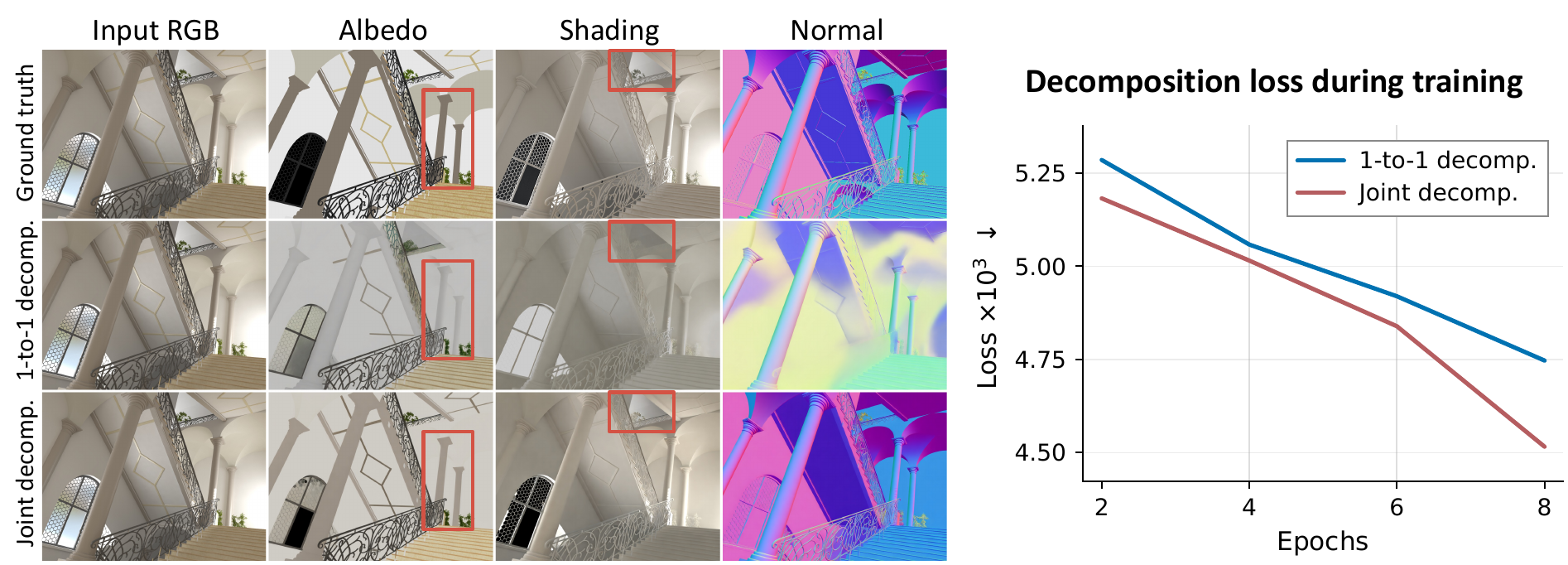} \caption{\textbf{Ablation on joint-channel decomposition.}
Compared with 1-to-1 decomposition, joint decomposition produces cleaner, more coherent intrinsic maps and consistently achieves lower training loss, demonstrating the benefit of cross-channel interaction.}
\label{fig:ablation_sep_joint} 
\end{figure}

\begin{figure*}[!t] 
\centering \includegraphics[width=\textwidth]{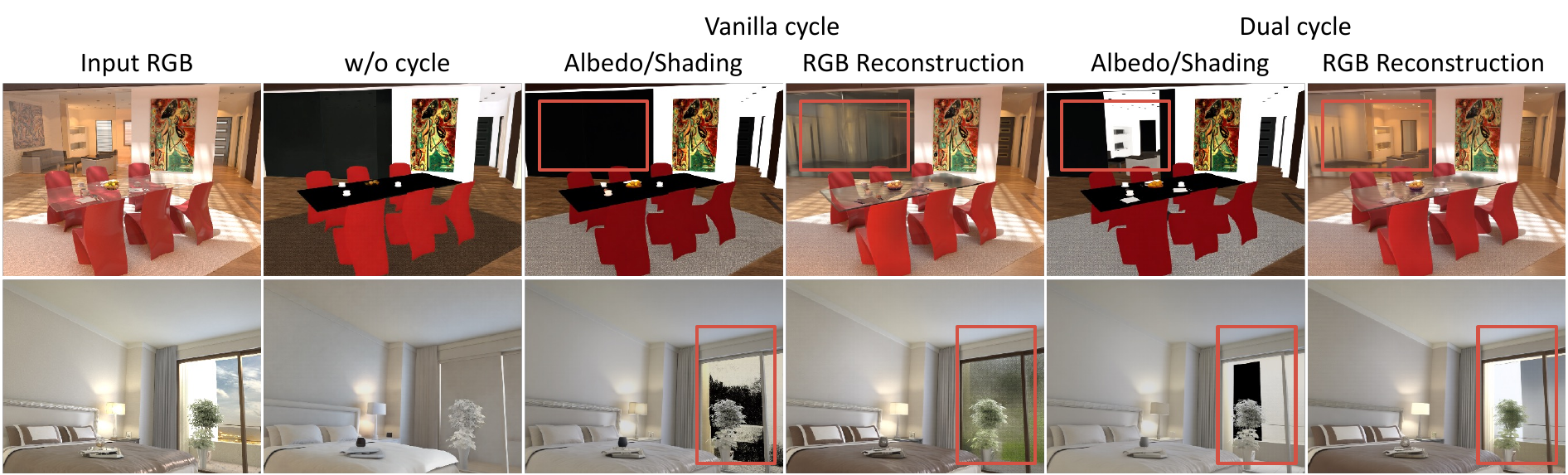} \caption{\textbf{Ablation on cycle consistency.}
Compared with the vanilla cycle, our dual cycle better preserves the semantic roles of intrinsic channels while maintaining faithful RGB reconstruction.}
\label{fig:ablation_cycle} 
\end{figure*}

\paragraph{Joint-channel decomposition}
We first examine the effect of joint-channel decomposition by comparing the 1-to-1 decomposition baseline (a) with the joint variant (c) in \Cref{tab:ablation}. Neither of them uses the cycle constraint. Joint prediction improves all six reported decomposition metrics: albedo and irradiance PSNR improve by 1.27\,dB and 0.91\,dB, respectively, while the mean angular error for normals decreases by $2.32^\circ$. These consistent improvements across appearance, illumination, and geometry support the benefit of jointly modeling intrinsic channels over per-channel generation. \Cref{fig:ablation_sep_joint} further shows that joint decomposition achieves cleaner intrinsic maps and lower training loss, as cross-channel interaction provides complementary cues to reduce the ambiguity of individual prediction.

\paragraph{Cycle Consistency}
We further compare joint decomposition without cycle supervision (c), with a vanilla cycle loss (d), and with our dual cycle loss (e) in Table~\ref{tab:ablation}. Introducing the vanilla cycle improves normal estimation, perceptual quality, and RGB reconstruction, while reducing the PSNR of albedo and irradiance compared with (c). We attribute this result to a shortcut introduced by the input-reconstruction supervision.
To cycle back to the original RGB image, the model may leak information across intrinsic channels. Such leakage can benefit reconstruction without contributing to a better physical decomposition.
In contrast, our dual cycle further achieves better albedo and irradiance PSNR, normal accuracy, and forward-rendering performance.

\subsection{Applications}
\begin{figure}[!t] 
\centering \includegraphics[width=\linewidth]{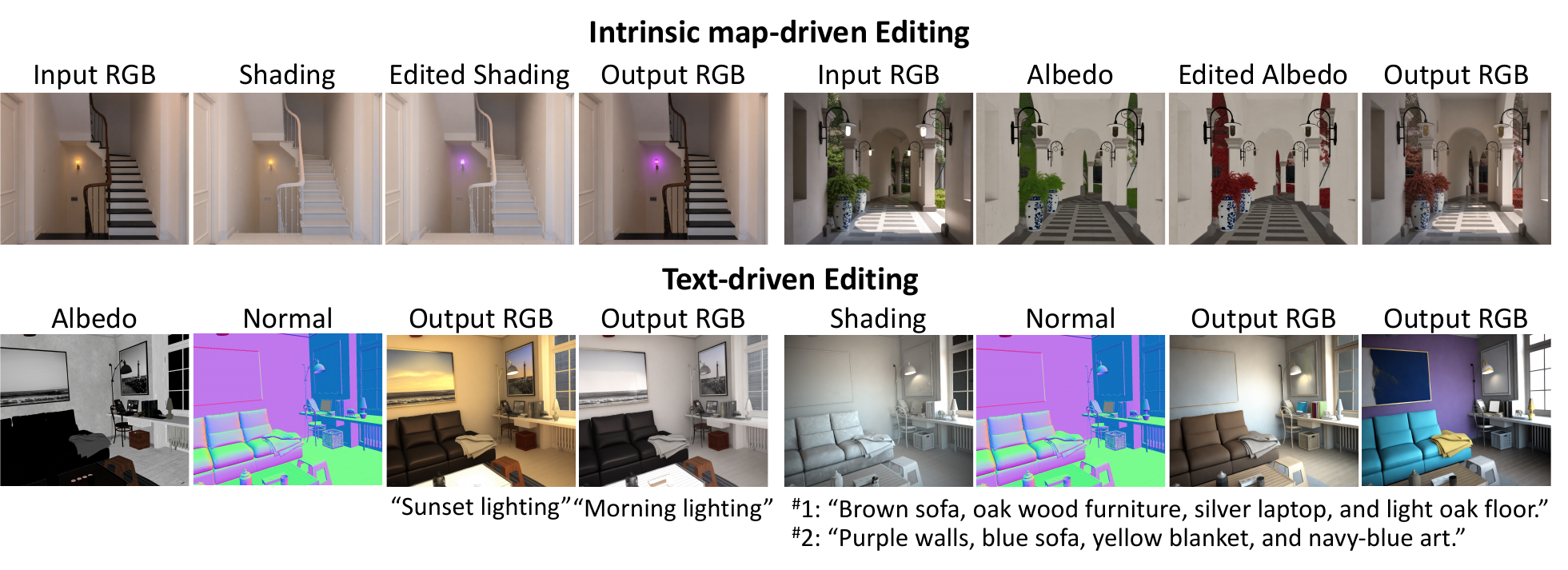} \caption{\textbf{Intrinsic-guided image editing.} Our framework enables controllable image editing by either directly manipulating intrinsic channels or using text prompts to modify selected properties.} 
\label{fig:conditional_rendering} 
\end{figure}

Our reciprocal framework supports controllable image editing through the intermediate intrinsic representation.
As shown in \Cref{fig:conditional_rendering}, we explore two types of intrinsic-guided image editing.
First, the decomposed intrinsic maps can be edited individually and then rendered back to RGB.
This enables explicit control over scene properties such as illumination and surface materials.
Second, our forward renderer can synthesize images from a subset of intrinsic channels, with text prompts specifying the unconstrained properties.
These results demonstrate the flexibility of \method{} for fine-grained image manipulation through explicit control of physical scene properties. Additional results are shown in the appendix.

\section{Conclusion, Limitations and Future Work}

In this work, we presented \method, a unified framework for joint intrinsic decomposition and reciprocal forward rendering. 
Rather than treating intrinsic decomposition and rendering as separate tasks, our framework models them as two reciprocal views of the same physical image formation process. 
\gz{Our results suggest that jointly reasoning across intrinsic channels while coupling decomposition with forward rendering provides an effective way to promote physically coherent scene representations, with improved interdependence across intrinsic channels.}
\zs{This interdependence operates at two levels: cross-channel interaction jointly models intrinsic properties, and cross-process interaction links inverse and forward rendering. Together, they reduce feature leakage and encourage decompositions that are accurate at the channel level and consistent at the \gz{process} level.}

Our current implementation still has several limitations. 
First, inference remains relatively expensive, requiring approximately 90 seconds to decompose a single $1024 \times 768$ image into five intrinsic channels using 30 denoising steps. 
\gz{Second, available intrinsic datasets vary substantially in their channel availability and scene distributions, limiting the completeness and diversity of physical supervision.
Going further, the scarcity of accurate intrinsic annotations for real scenes still leaves a sim-to-real gap for complex geometry and illumination.
Our future work includes exploring more efficient sampling and larger-scale training to further improve the efficiency and generalization.}

\bibliographystyle{iclr2027_conference}
\bibliography{reference}

\clearpage
\appendix

\section*{\textbf{Appendix}}
\section{Implementation Details}

As shown in \Cref{tab:datasets}, our training set comprises about 100K images collected from three synthetic datasets and four real-image datasets. Since the real-image datasets do not provide ground-truth intrinsic annotations, we employ Ouroboros to provide pseudo labels for supervision. 
During training, dataset \(i\) is sampled with probability: $p_i = \frac{\sqrt{N_i}}{\sum_j \sqrt{N_j}}$,
where \(N_i\) denotes the number of training samples in dataset \(i\). We use a fixed ordering for both input and output slots: albedo, shading, normal, roughness, metallic. During the decomposition stage, missing intrinsic channels were set to zero-valued latents and masked out from loss calculation. During the composition stage, each available intrinsic condition is independently dropped out with a probability of $0.05$.
The direct decomposition and composition losses both have a weight of $1.0$, while the decomposition loss on real images is weighted by $0.5$. 
For the cycle consistency loss, we average the losses from the two directions and linearly increase the weight from $0$ to $0.5$ in the first 1,000 steps.
We use separate LoRA adapters for decomposition and composition while sharing the pretrained MMDiT backbone between the two directions.
The LoRA rank and alpha are set to 32. 
Training is performed on 5 NVIDIA Pro6000 GPUs for 35K iterations and takes about 72 hours.


\begin{table}[h]
\centering
\caption{\textbf{Summary of training dataset.}}
\label{tab:datasets}
\resizebox{\textwidth}{!}{
\begin{tabular}{lcccc}
\toprule
\textbf{Dataset} &
\textbf{\# of Images} &
\textbf{Resolution} &
\textbf{Scene Type} &
\textbf{Ground truth Intrinsic Types} \\
\midrule

Hypersim~\citep{roberts2021hypersim}
& 20,000
& $1024 \times 768$
& Synthetic, Indoor
& Albedo, Shading, Normal \\

InteriorVerse~\citep{zhu2022learning}
& 15,000
& $640 \times 480$
& Synthetic, Indoor
& Albedo, Normal, Roughness, Metallic \\

MatrixCity~\citep{li2023matrixcity}
& 20,000
& $768 \times 768$
& Synthetic, Outdoor
& Albedo, Normal, Roughness, Metallic \\

MIDIntrinsics~\citep{murmann2019multi}
& 10,000
& $768 \times 768$
& Real, Indoor
& - \\

DL3DV\_960p~\citep{ling2024dl3dv}
& 12,800
& $768 \times 432$
& Real, Indoor
& - \\

ScanNet++~\citep{yeshwanth2023scannetpp}
& 20,000
& $512 \times 512$
& Real, Indoor
& - \\

Adobe FiveK~\citep{vladimir2011fivek}
& 1,453
& $X \times 768$
& Real, Outdoor
& - \\

\bottomrule
\end{tabular}
}
\end{table}

\section{Additional Results}

We provide additional results to further evaluate generalization and controllability of our method. We first report comparisons on synthetic datasets with ground-truth in \Cref{synthetic_comparison}. We then assess its performance on real-world scenes using indoor and outdoor images from the test dataset, as well as out-of-distribution (OOD) images collected from publicly available photography websites in \Cref{real_comparison}. Finally, we demonstrate intrinsic-guided image editing by manipulating individual intrinsic channels while keeping the remaining components fixed in \Cref{image_editing}.

\subsection{Comparison on Synthetic Images}
\label{synthetic_comparison}

\begin{figure*}[htbp]
\centering 
\includegraphics[width=0.98\linewidth]{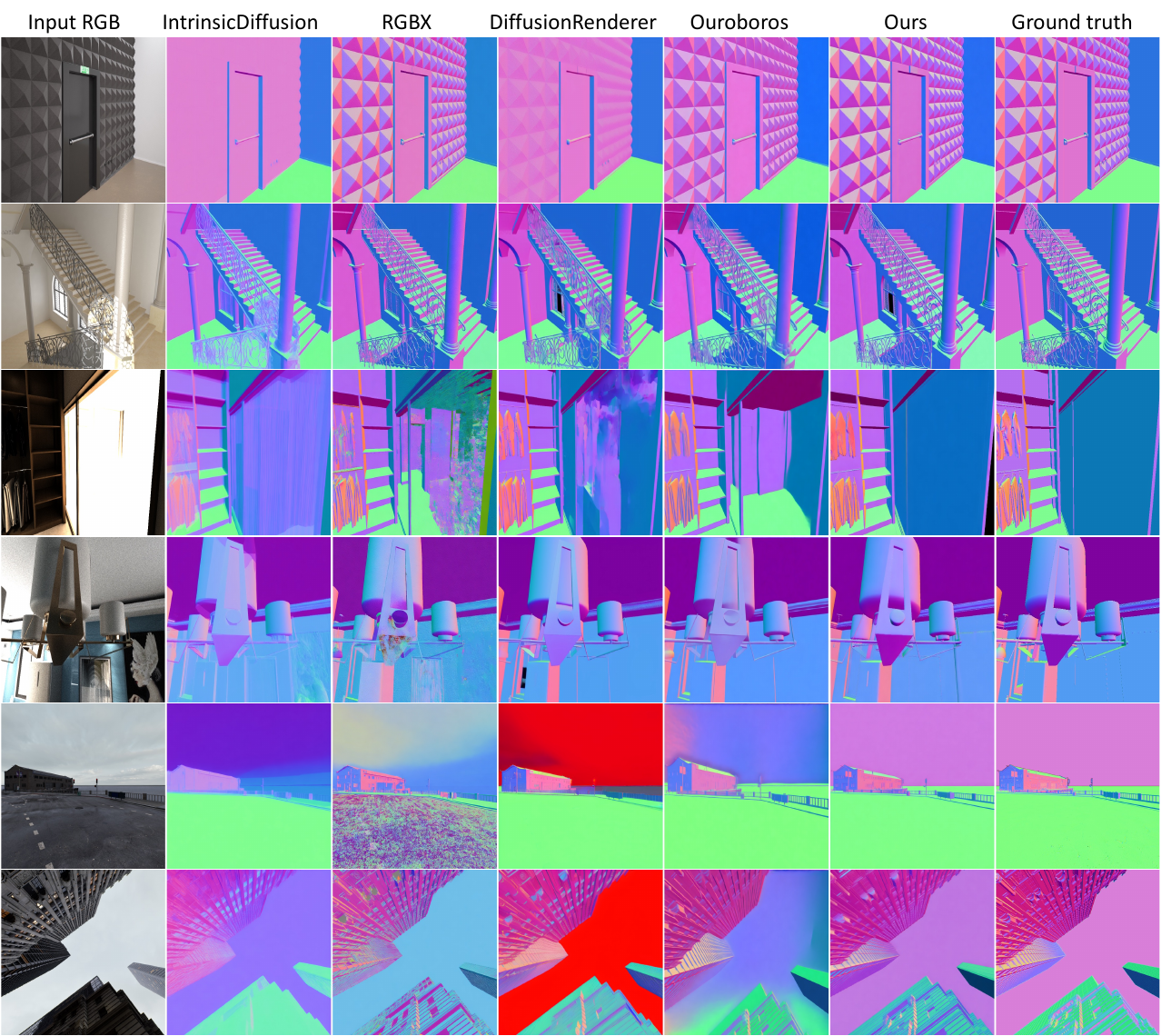} 
\caption{\textbf{{Qualitative comparison of normal estimation on synthetic images.}}} 
\label{fig:comparison_normal_sync_data} 
\end{figure*}

\begin{figure*}[htbp]
\centering 
\includegraphics[width=0.98\linewidth]{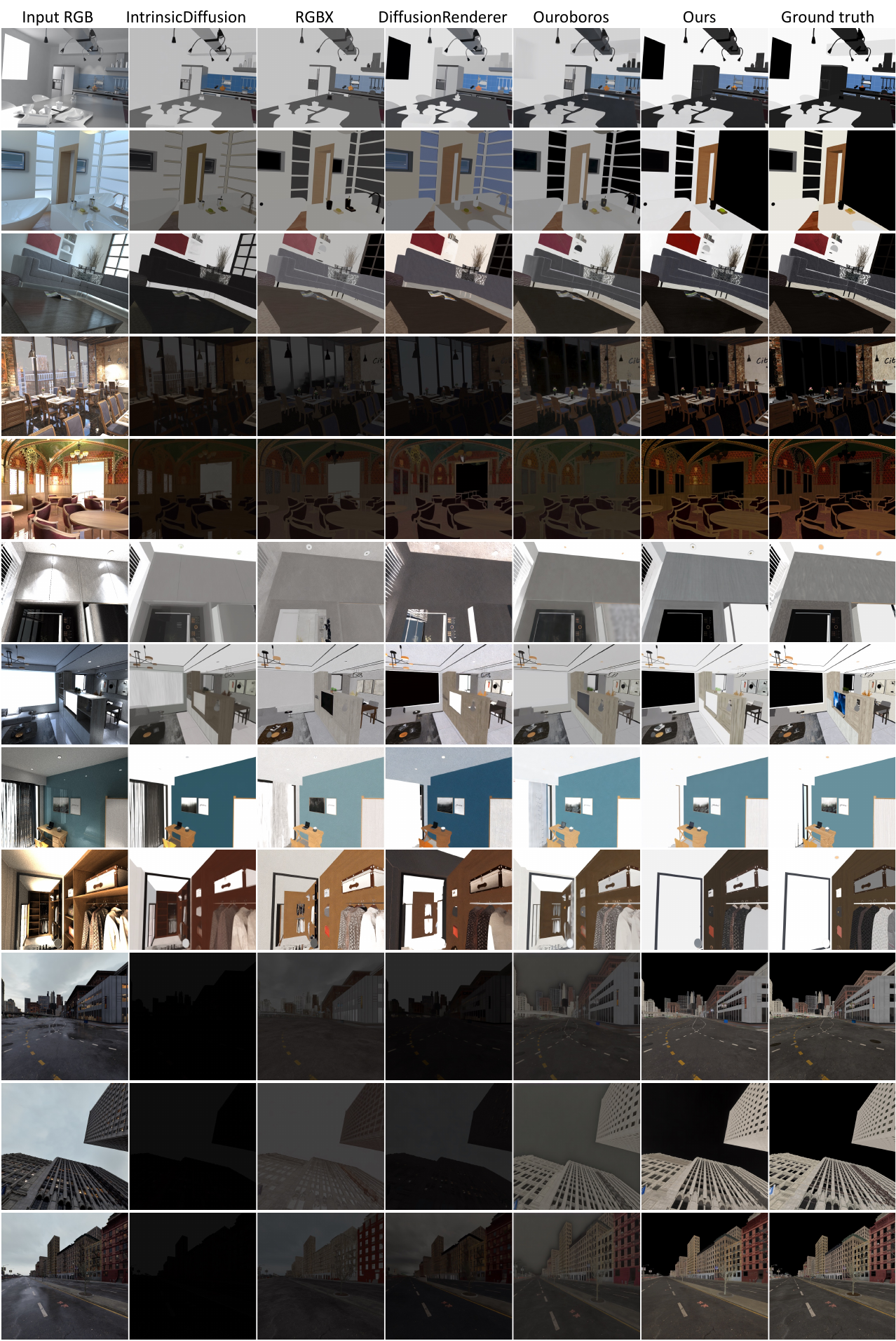} 
\caption{\textbf{Qualitative comparison of albedo estimation on synthetic images.}
We show additional results across diverse synthetic scenes, comparing the predicted albedo maps of our method with existing approaches and the corresponding ground truth.} 
\label{fig:comparison_albedo_sync_data} 
\end{figure*}

\begin{figure*}[htbp]
\centering 
\includegraphics[width=0.98\linewidth]{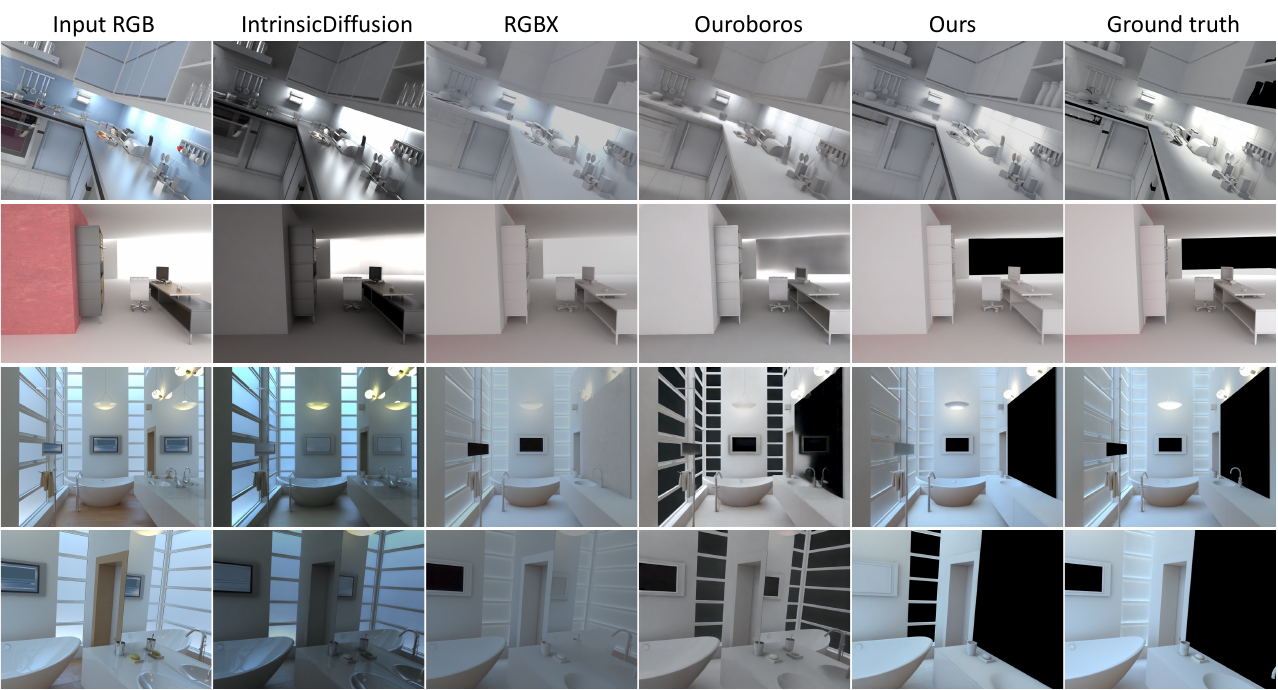} 
\caption{\textbf{{Qualitative comparison of shading estimation on synthetic images.}}} 
\label{fig:comparison_shading_sync_data} 
\end{figure*}

\begin{figure*}[htbp]
\centering 
\includegraphics[width=0.98\linewidth]{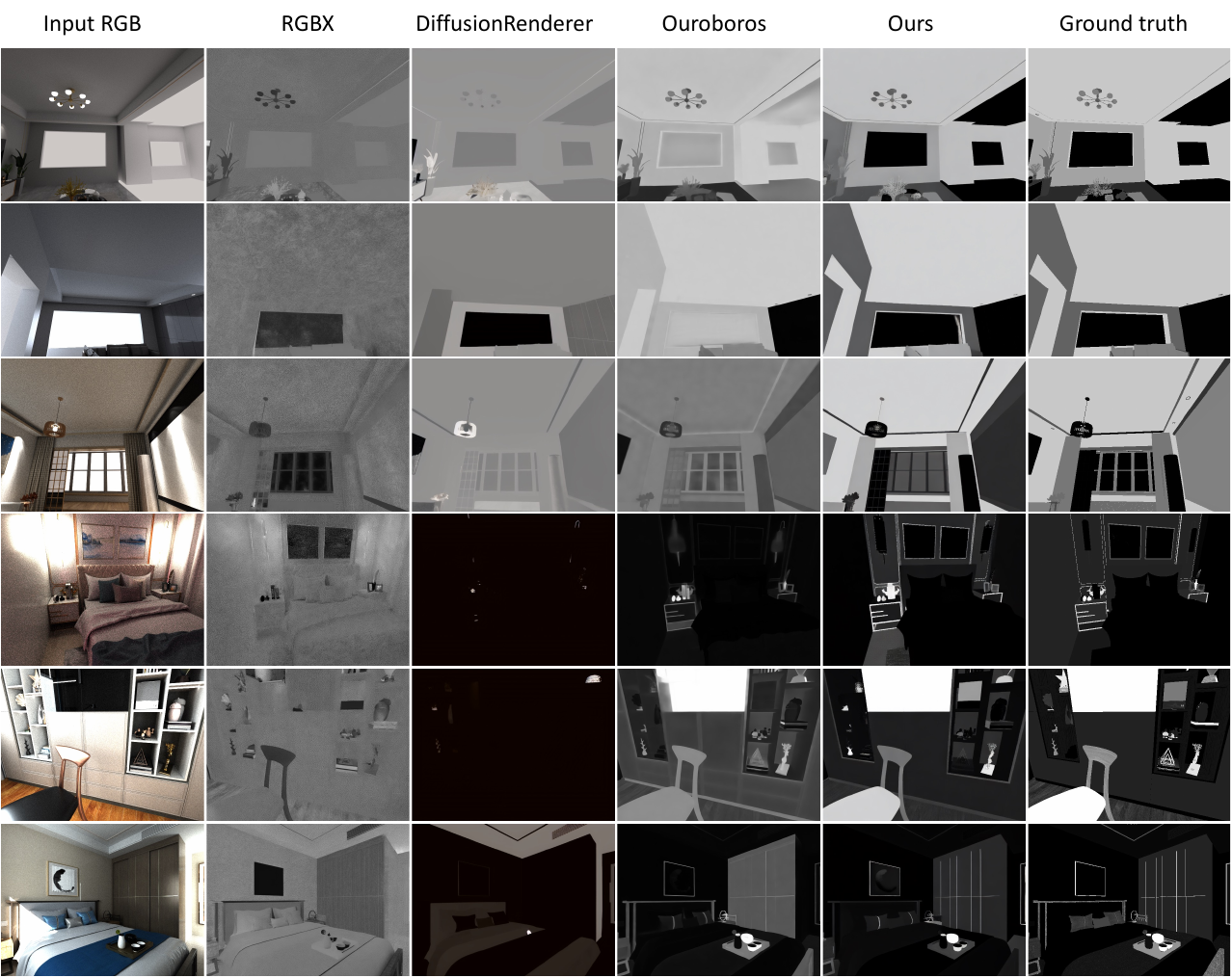} 
\caption{\textbf{{Qualitative comparison of roughness and metallic estimation on synthetic images.}}} 
\label{fig:comparison_r_m_sync_data} 
\end{figure*}

We present additional comparisons on the test sets of Hypersim, InteriorVerse, and MatrixCity in
\Cref{fig:comparison_albedo_sync_data,fig:comparison_shading_sync_data,fig:comparison_normal_sync_data,fig:comparison_r_m_sync_data}.
Since the available annotations vary across datasets, each method is
evaluated only on the intrinsic components for which ground-truth annotations
are available. Overall, our method achieves the best decomposition quality
across the three datasets, producing intrinsic maps that most closely match
the ground truth in both structure and appearance. In particular, it more
effectively disentangles reflectance from illumination in the albedo and
shading components, preserves geometric details in the predicted normals,
and recovers more accurate and spatially coherent roughness and metallicity.

\subsection{Comparison on Real Images}
\label{real_comparison}

\begin{figure*}[htbp]
\centering 
\includegraphics[width=0.98\linewidth]{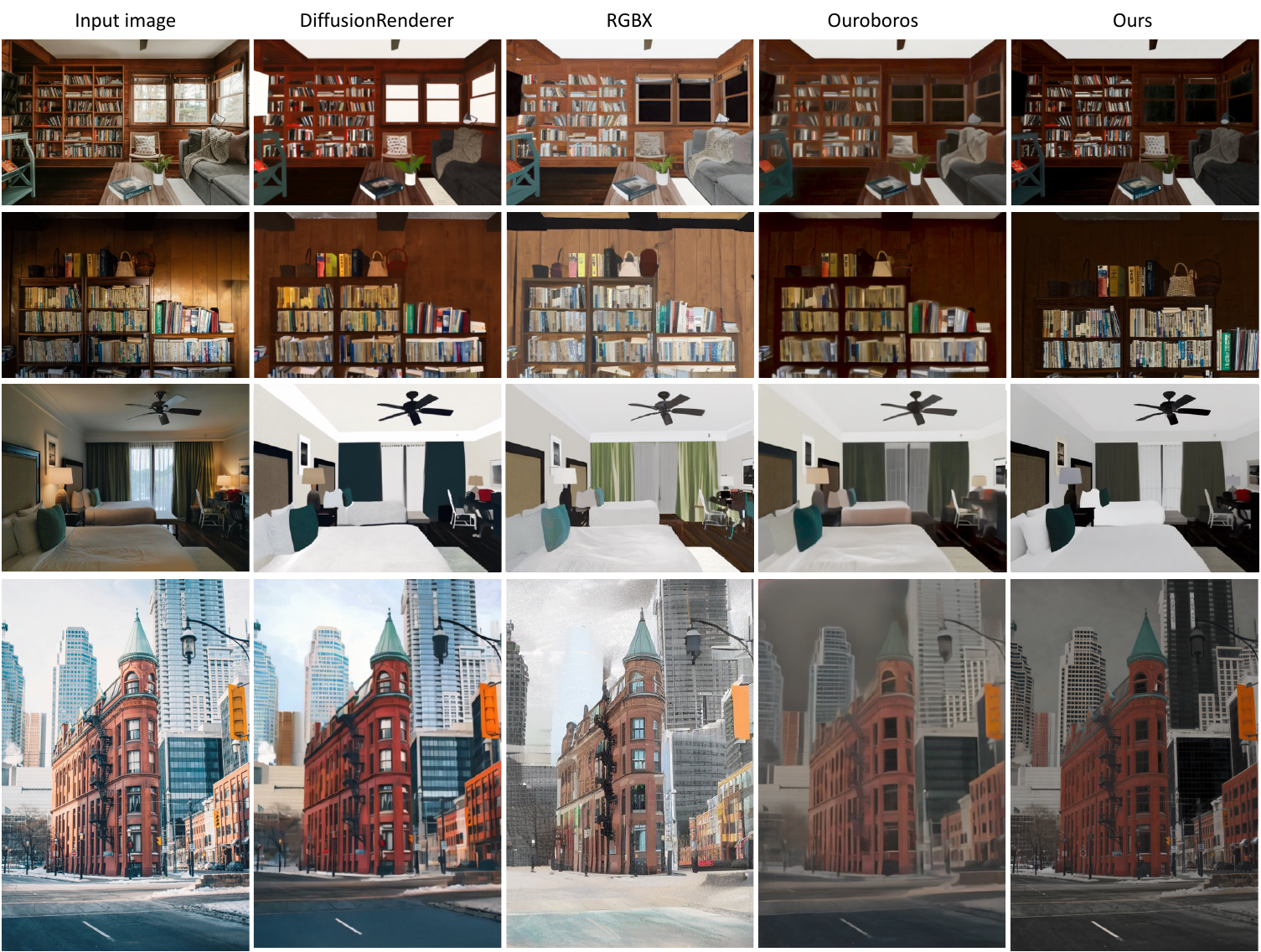} 
\caption{\textbf{{Qualitative comparison of albedo estimation on real images.}}} 
\label{fig:comparison_albedo_real_image}
\end{figure*}

\begin{figure*}[htbp]
\centering 
\includegraphics[width=0.98\linewidth]{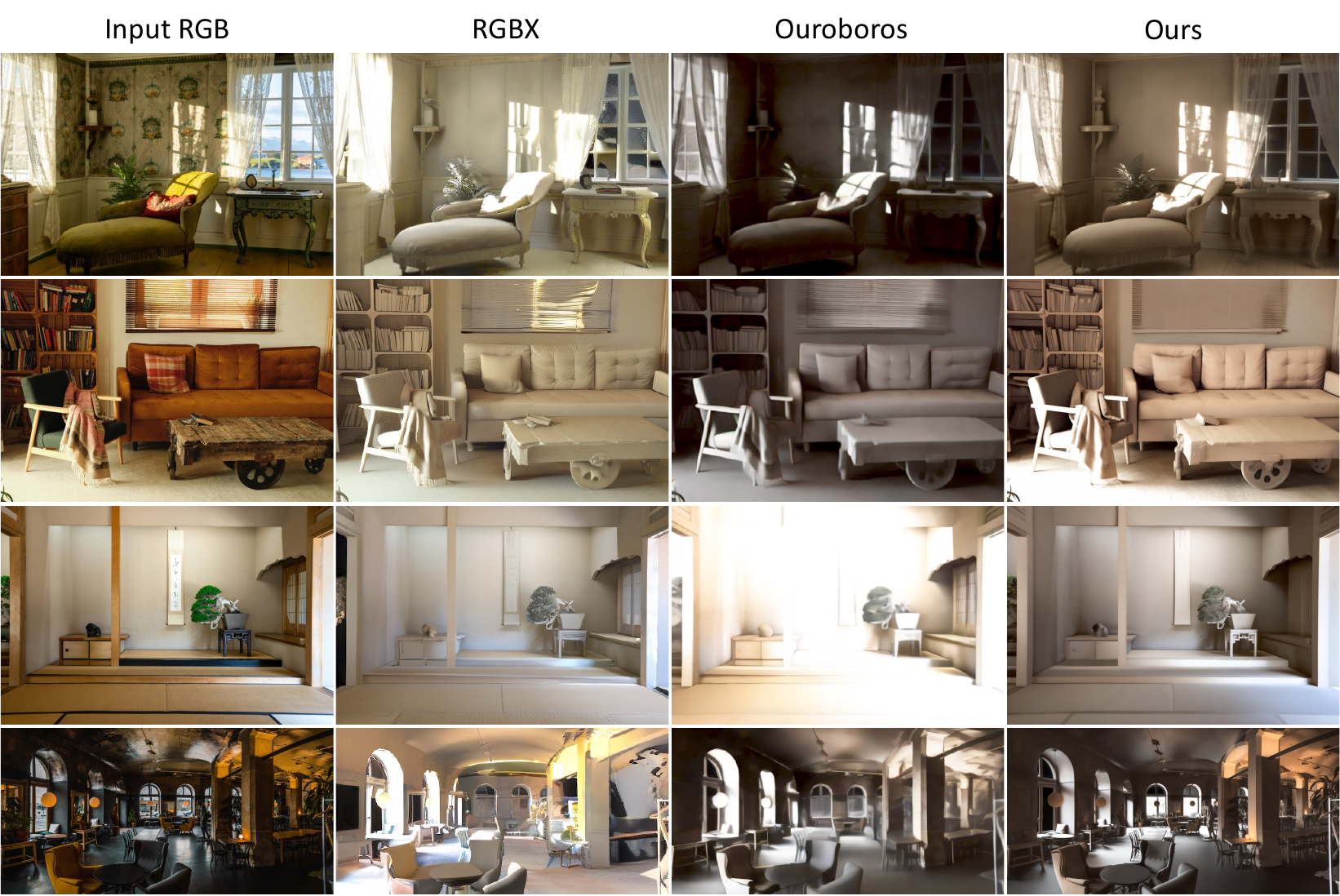} 
\caption{\textbf{Qualitative comparison of shading estimation on real images.}} 
\label{fig:comparison_shading_real_image}
\end{figure*}

\begin{figure*}[htbp]
\centering 
\includegraphics[width=0.98\linewidth]{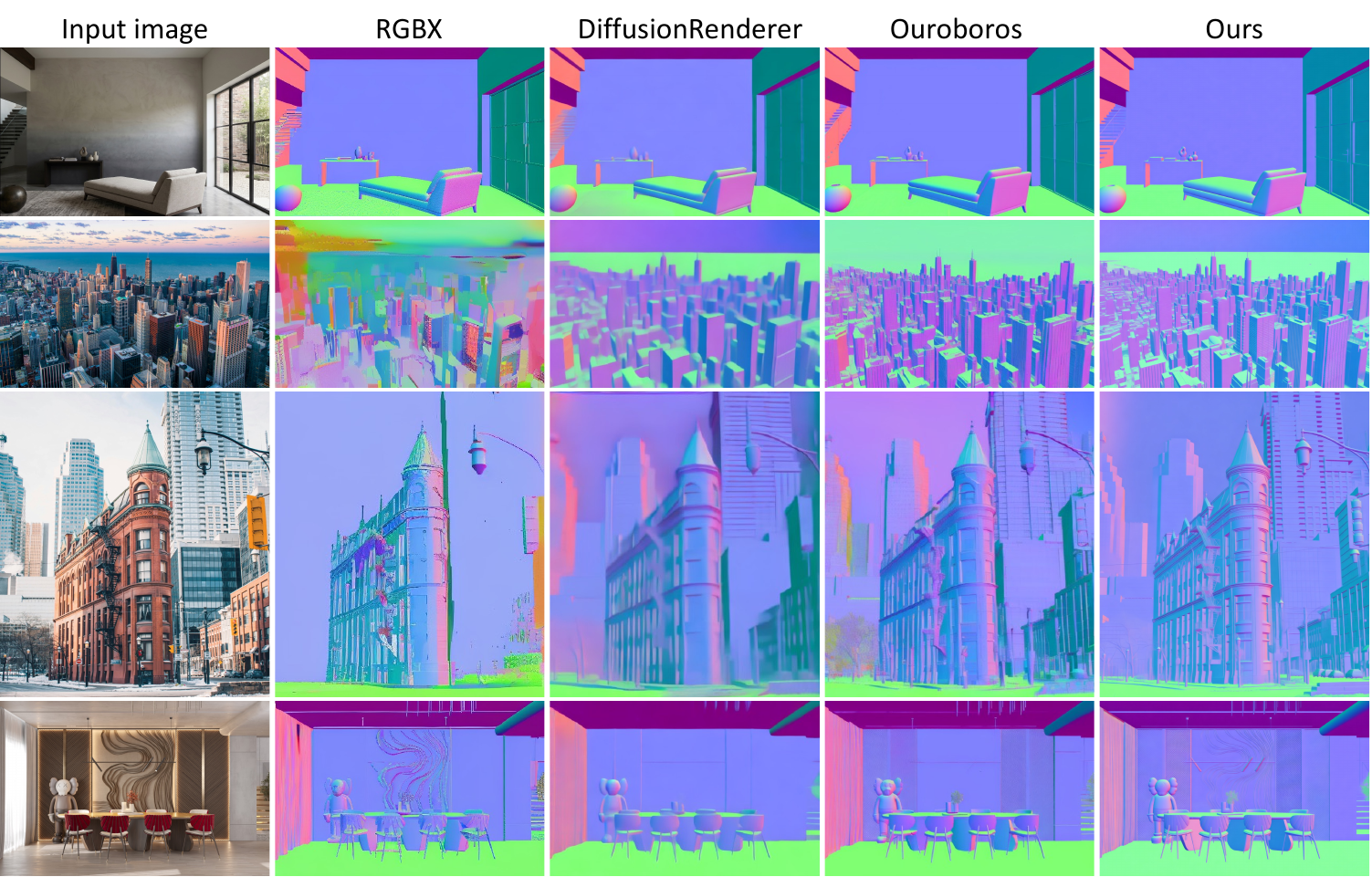} 
\caption{\textbf{{Qualitative comparison of normal estimation on real images.}}} 
\label{fig:comparison_normal_real_image} 
\end{figure*}

\begin{figure*}[htbp]
\centering 
\includegraphics[width=0.98\linewidth]{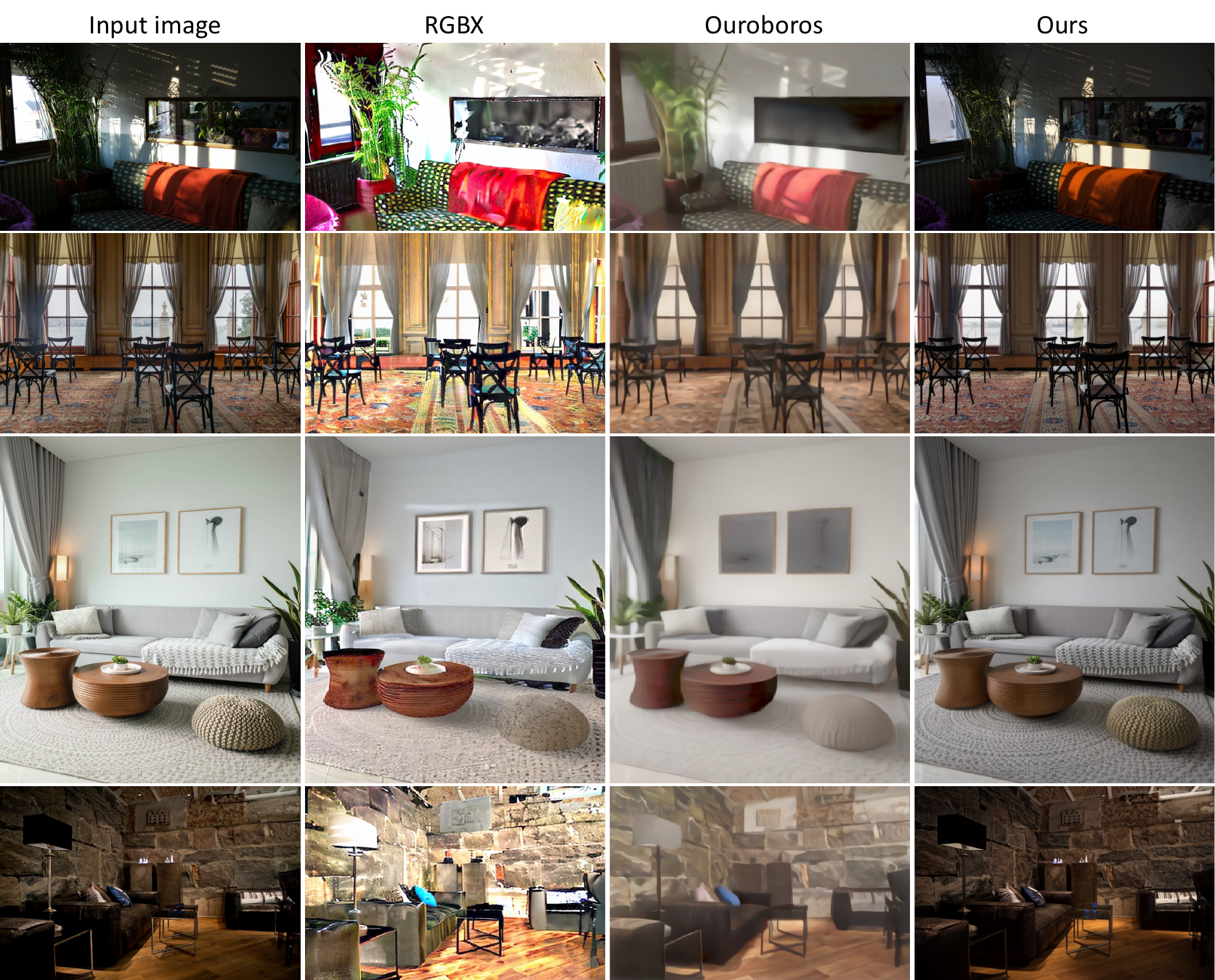} 
\caption{\textbf{{Qualitative comparison of RGB reconstruction on real images.}}} 
\label{fig:comparison_recon_real_image} 
\end{figure*}

We further compare the methods on real-world indoor and outdoor images in
\Cref{fig:comparison_albedo_real_image,fig:comparison_shading_real_image,fig:comparison_normal_real_image,fig:comparison_recon_real_image}.
Because ground-truth intrinsic annotations are unavailable for those images, we focus on their visual plausibility, consistency with the given inputs and cross-intrinsic leakage. All methods are evaluated under the same input images and text prompts. Compared with existing methods, our approach produces more visually plausible results while better preserving the content and structure of the input images. Moreover, the predicted intrinsic components exhibit less cross-intrinsic leakage, indicating improved \gz{interdependence among} different intrinsic properties.

\subsection{Intrinsic-guided Image Editing}
\label{image_editing}
Beyond intrinsic decomposition and RGB reconstruction, we examine whether the learned intrinsic representations provide continuous and interpretable control over the reconstruction results. Given an input image, we first decompose it into five intrinsic channels and then independently manipulate either albedo or shading component while keeping the remaining components and text condition fixed. Specifically, we vary the albedo saturation using factors of \(1.0, 1.5, 2.0,\) and \(3.0\). For shading, the brightness factors are \(0.25, 0.5, 1.0, 1.5,\) and \(2.0\).

\begin{figure*}[t]
\centering 
\includegraphics[width=\textwidth]{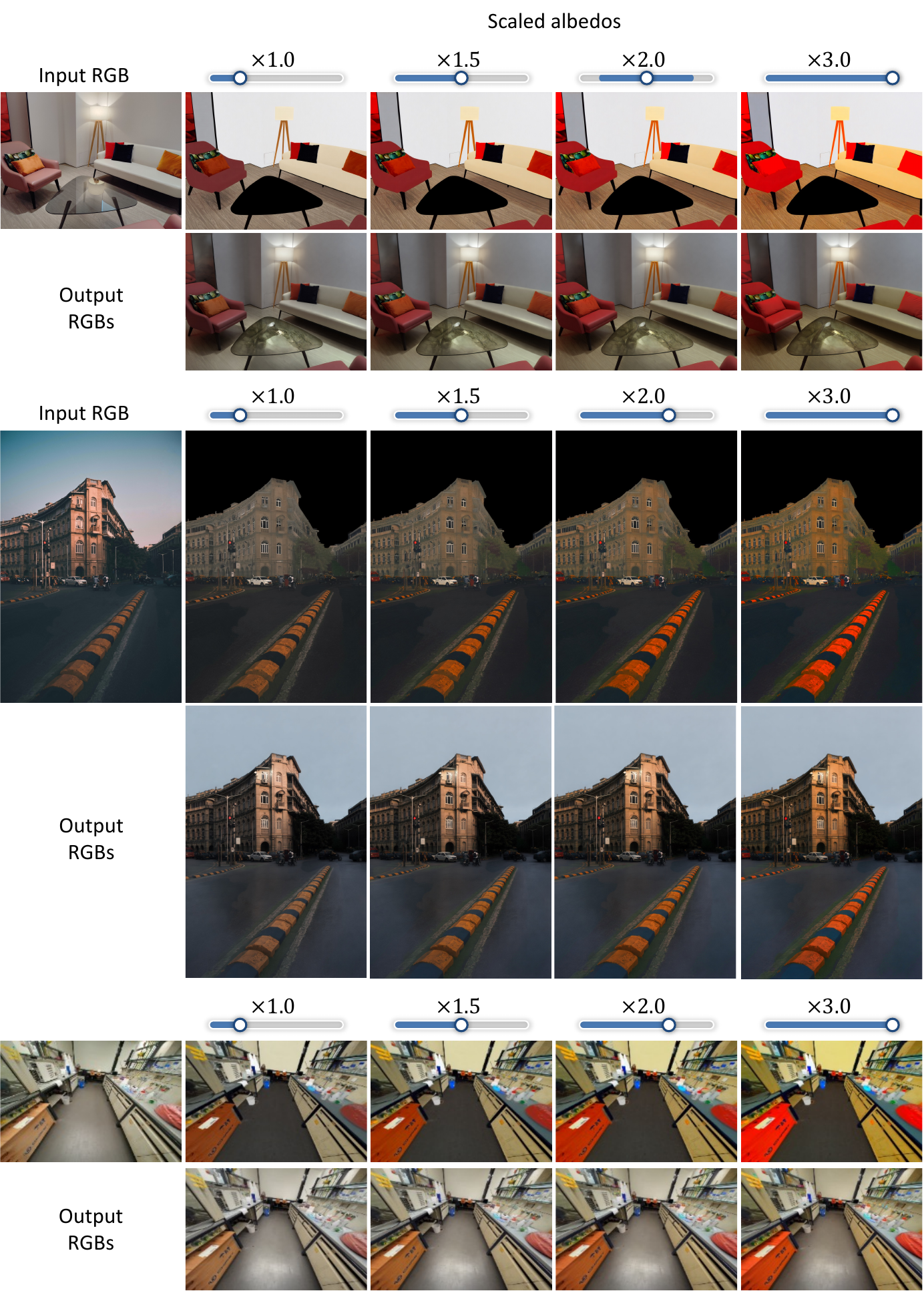} 
\caption{\textbf{Results of global albedo scaling.}
We show additional examples of globally scaling the decomposed albedo and rendering the modified intrinsic maps back to RGB, demonstrating controllable changes in surface appearance.}
\label{fig:global_albedo_scaling} 
\end{figure*}

\begin{figure*}[t]
\centering 
\includegraphics[width=\textwidth]{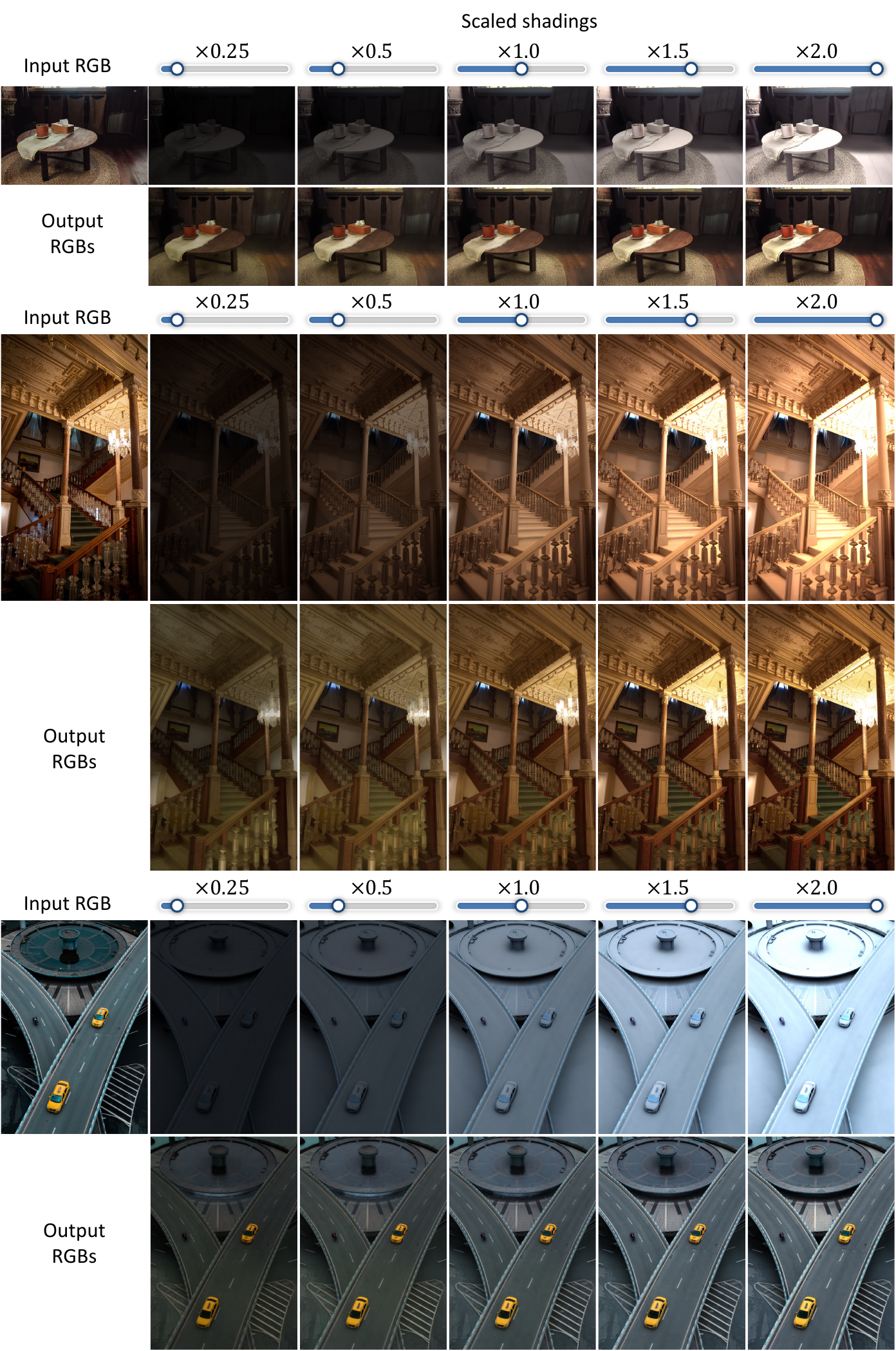} 
\caption{\textbf{Results of global relighting.}
We show additional examples of global relighting, where the illumination is modified while preserving the underlying scene content and intrinsic properties.}
\label{fig:global_relighting} 
\end{figure*}

As shown in 
\Cref{fig:global_albedo_scaling,fig:global_relighting},
increasing or decreasing the albedo
saturation produces smooth and corresponding changes in surface colorfulness, while largely preserving the original illumination and scene structure. Similarly, adjusting the shading brightness continuously changes the perceived illumination intensity without introducing noticeable changes to surface reflectance or scene content. These results demonstrate that the learned intrinsic representations enable smooth and interpretable control while maintaining a clear separation between albedo and shading.

\begin{figure*}[t]
\centering 
\includegraphics[width=\textwidth]{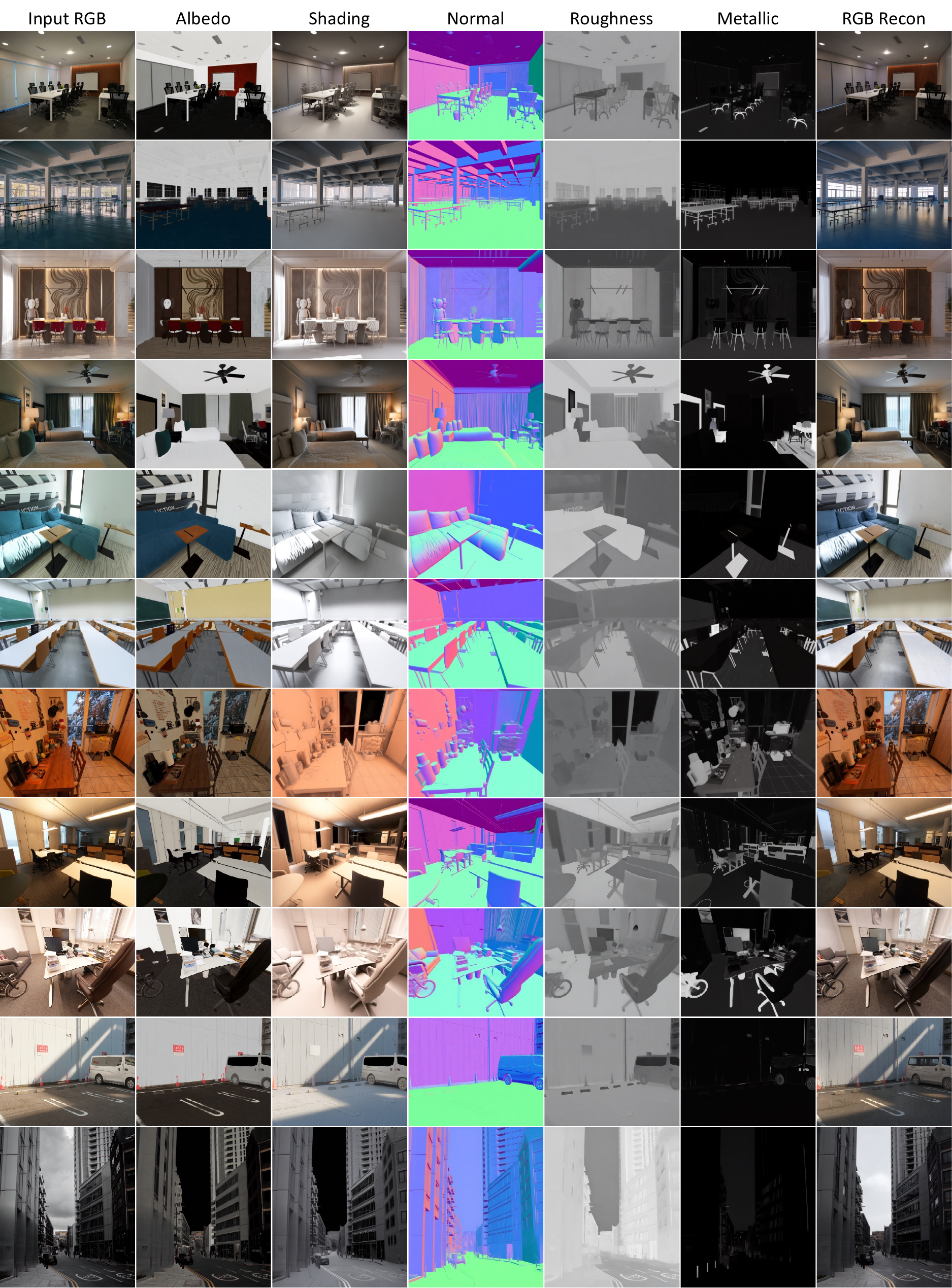} 
\caption{ \textbf{Results of Inverse and forward rendering on real images.} We evaluate our method on diverse real-world scenes, demonstrating its ability to jointly estimate clean and physically consistent intrinsic channels and rendering the input RGB images.} 
\label{fig:real_image_decomp_ours} 
\end{figure*}

\end{document}